\documentclass[runningheads]{llncs}

\usepackage{eccv}

\usepackage{eccvabbrv}
\usepackage{booktabs}
\usepackage{graphicx}
\usepackage{booktabs}
\usepackage{graphicx}
\usepackage{amsmath}
\usepackage{amssymb}
\usepackage{booktabs}
\usepackage{caption}
\usepackage[normalem]{ulem}
\usepackage{tabularray}
\usepackage{marvosym}
\usepackage{multirow}
\usepackage{diagbox}
\usepackage{makecell}
\usepackage{colortbl}
\usepackage{color}
\usepackage{subfloat}
\usepackage[marginal]{footmisc}
\usepackage{ulem}
\usepackage[accsupp]{axessibility}  
\newenvironment{tablefont}{\fontfamily{ptm}\selectfont}

\usepackage[pagebackref,breaklinks,colorlinks,citecolor=eccvblue]{hyperref}
\usepackage{orcidlink}

\begin{document}


\title{Norface: Improving Facial Expression Analysis by Identity Normalization}
\titlerunning{Norface}


\author{Hanwei Liu\inst{1,2,5}$^*$\orcidlink{0000-0003-0568-8708} \and
Rudong An\inst{2}$^*$\orcidlink{0000-0002-8575-8229} \and
Zhimeng Zhang\inst{2}\orcidlink{0000-0002-3695-1129} \and
Bowen Ma\inst{2,5}\orcidlink{0000-0002-7538-3996} \and
Wei Zhang\inst{2}\orcidlink{0000-0001-5907-7342} \and
Yan Song\inst{2}\orcidlink{0009-0006-0142-4915} \and
Yujing Hu\inst{2}\orcidlink{0000-0002-2714-0092} \and
Wei Chen\inst{3} \and
Yu Ding\inst{2,4,5}$^*$\orcidlink{0000-0003-1834-4429}
}

\renewcommand{\thefootnote}{\fnsymbol{footnote}}
\footnotetext[1]{Equal contribution. Yu Ding is the corresponding author.}

\authorrunning{H Liu. et al.}


\institute{$^1$Tongji University \quad 
$^2$Netease Fuxi AI Lab \quad  
$^3$Hebei Agricultural University   \\
$^4$School of Computer Science, Hangzhou Dianzi University \quad  
$^5$Happy Elements}


\maketitle

\begin{center}
    \includegraphics[width=0.95\linewidth]{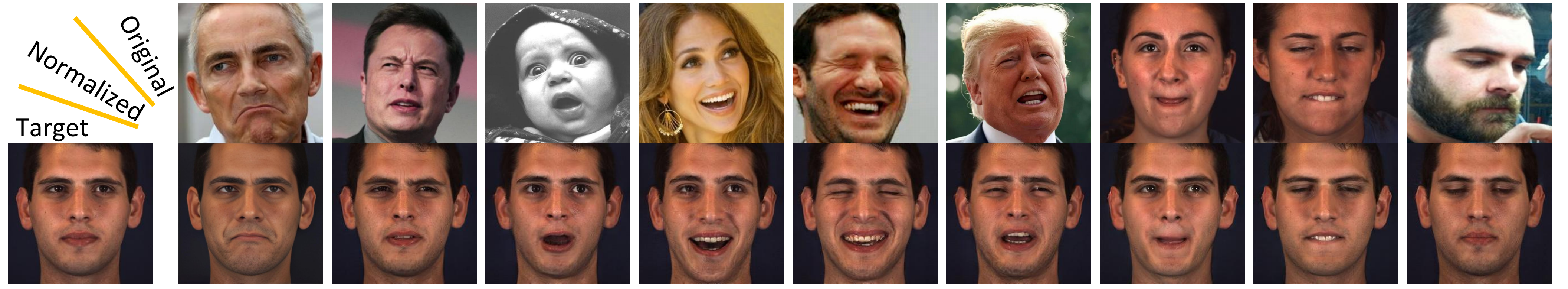}
    \captionof{figure}{Identity normalization. The figure demonstrates that various original images (i.e. in-the-lab and in-the-wild ones) with AUs and emotions are normalized to a common target identity with a consistent pose, background, etc.} 
\label{fig_intro} 
\end{center}

\begin{abstract}
Facial Expression Analysis remains a challenging task due to unexpected task-irrelevant noise, such as identity, head pose, and background. To address this issue, this paper proposes a novel framework, called Norface, that is unified for both Action Unit (AU) analysis and Facial Emotion Recognition (FER) tasks. Norface consists of a normalization network and a classification network. First, the carefully designed normalization network struggles to directly remove the above task-irrelevant noise, by maintaining facial expression consistency but normalizing all original images to a common identity with consistent pose, and background. Then, these additional normalized images are fed into the classification network. Due to consistent identity and other factors (e.g. head pose, background, etc.), the normalized images enable the classification network to extract useful expression information more effectively. Additionally, the classification network incorporates a Mixture of Experts to refine the latent representation, including handling the input of facial representations and the output of multiple (AU or emotion) labels. Extensive experiments validate the carefully designed framework with the insight of identity normalization. The proposed method outperforms existing SOTA methods in multiple facial expression analysis tasks, including AU detection, AU intensity estimation, and FER tasks, as well as their cross-dataset tasks. For the normalized datasets and code please visit \href{https://norface-fea.github.io/}{project page}.

\end{abstract}

\section{Introduction}
\label{sec:intro}
Facial Expression Analysis (FEA) is a complex task in affective computing that often involves facial Action Units (AU) analysis (AU detection and AU intensity estimation) and Facial Emotions Recognition (FER). Despite the progress made so far \cite{ref3,ref8, ref23, ref60, ref61,ref135}, both FEA tasks still face challenges due to the inherent entanglement of expressions with unexpected factors like identity, head pose, and background \cite{ref1_1,ref1_2}.

Specifically, FEA suffers from severe identity bias \cite{ref103,ref1_1}, which exacerbates the intricacies of refinement and representation of facial expressions \cite{ref91,ref98}. FEA model struggles to generalize effectively to unseen identities, this bias also leads the model to overfit seen identities and undermines the overall performance \cite{ref92,ref99}. In addition, variations in pose and background compound the challenges in FEA tasks \cite{ref3,ref104}. Hence, a crucial objective of FEA is to mitigate interference arising from irrelevant factors, including identity, pose, and background.


To address these challenges, some studies attempt to construct identity-based expression pairs \cite{ref98,ref99,ref100,ref101,ref102} to obtain identity-invariant representations. Due to significant variations in identities, the issue of identity bias still persists. Alternatively, certain studies aim to incorporate generative techniques, such as generating 'neutral' emotions \cite{ref91,ref92,ref93}, 'average' identities \cite{ref94}, or diverse emotions \cite{ref95,ref96,ref97}, to disentangle expression from identity. Nonetheless, these methods either suffer from limited generation quality or heavily rely on controlled lab datasets, thereby restricting their generalizability in complex and diverse real-world scenarios. 
Moreover, these approaches often overlook non-identity noise factors, such as pose and background variations. Additionally, their frameworks are often tailored to either AU analysis or FER tasks, which cannot be unified for both tasks, despite task-irrelevant noise being common to both.

To address the above challenges, this paper introduces a novel framework called \textbf{Norface}, which is applicable to AU detection,  AU intensity estimation, and FER tasks. The core concept behind Norface is \textbf{Identity normalization (Idn)}, which normalizes all images to a common identity with consistent pose and background, as shown in Fig.\ref{fig_intro}. By doing so, the resulting normalized data is intended to retain only the facial expression variations that are relevant to the task. Consequently, Idn mitigates the influence of identity bias, pose variations, and background changes, which are commonly encountered in both AU analysis and FER tasks. This process can also be interpreted as the removal of the above task-irrelevant noise, which is carried out by a normalization network in the first stage. Then, in the second stage, normalized images, as complementary to the original images, are used to improve expression analysis (AU detection, AU intensity estimation, or emotion recognition) by a classification network.

Specifically, in the first stage, the normalization network achieves Idn from all original faces to a target face. Thus, our normalization network is not only suitable for in-the-wild data but also capable of meeting the high-quality requirement for expression consistency. Specifically, we employ a pre-trained Masked AutoEncoder (MAE) to extract facial features, effectively capturing both expression and other attributes. Then, we develop an Expression Merging Module (EMM) to adaptively merge expression features from the original faces into the target face. Moreover, we introduce an expression loss and an eyebrow loss to enforce expression consistency between the normalized faces and the original faces. 

In the second stage, the classification network relies on Mixture of Experts (MoE). MoE aims to learn several expert sub-branches which are automatically trained to dynamically activate task-specific experts. This work proposes Input and Output MoE modules, where the input MoE module can refine latent facial representation, and the output MoE module can facilitate the detection, estimation, or recognition of multiple AU or emotion labels. 

Additionally, there are several multi-task methods \cite{ref12, ref87, ref88, ref13} that combine AU detection and FER tasks, requiring co-annotated AU and emotion labels, which are often unavailable in most datasets. Unlike these multi-task methods, our Norface offers a unified framework for these tasks individually, without the requirement for co-annotated AU and emotion labels, thus avoiding the issue of task imbalance \cite{ref89, ref90} being common in those multi-task methods.


Our work is the first to develop a unified framework to address noise issues in three tasks of AU detection, AU intensity estimation, and FER. Our Norface outperforms the existing results on the three tasks, as well as their cross-dataset tasks, showcasing its superior performance.
To sum up, ours makes the following contributions:
\begin{itemize}
\item This work provides a novel insight and releases a normalized dataset to remove the task-irrelevant noise by identity normalization that normalizes all expression images to a common identity with consistent pose and background.
\item This work proposes a new unified Facial Expression Analysis framework named Norface, designed for AU detection, AU intensity estimation, or FER tasks. This framework addresses identity normalization using Masked AutoEncoder and representation refinement based on Mixture of Experts.
\item Extensive quantitative and qualitative experiments validate the effectiveness of Norface, surpassing existing methods in AU detection, AU intensity estimation, and FER tasks, as well as their cross-dataset tasks. 
\end{itemize} 

\section{Related Work}

\noindent\textbf{Identity-Invariant FEA.} Identity bias has always been a challenge for FEA tasks, as expressions and identities are inherently entangled. Some methods \cite{ref98,ref99,ref100,ref101,ref102} construct pairs of identity-based expressions to obtain identity-invariant representations. DLN \cite{ref100} and GLEE-Net \cite{ref62} construct triples mapping identity to facial expression similarity for performing FER and AU detection tasks, respectively. 
IdenNet \cite{ref102} leverages face clustering to extract identity-dependent features and perform AU detection. 
Moreover, synthetic images \cite{ref91,ref92,ref93,ref94,ref95,ref96,ref97} have been employed to decouple identity and expression. DeRL \cite{ref91} generates neutral facial images to represent lab facial images. Huang et al. \cite{ref95} synthesize basic emotions for each identity. Differing from these approaches, ours focuses on identity normalization to address identity bias.

\noindent\textbf{AU analysis and FER tasks.} 
Current AU analysis and FER methods often rely on auxiliary information to mitigate the interference of task-irrelevant noise, such as facial landmarks \cite{ref8,ref18,ref13,ref5}, emotional priors \cite{ref78,ref80,ref11}, AU descriptions \cite{ref9,ref7,ref77,ref79}, or patch-based learning methods \cite{ref2,ref4,ref19,ref106,ref110}. SEV-Net \cite{ref7} utilizes both semantic descriptions and visual features of AU to enhance AU detection. LA-Net \cite{ref8} incorporates facial landmarks to augment expression features. However, they still encounter interference from such noise as they rarely explicitly eliminate task-irrelevant noise.

\noindent\textbf{Expression reenactment.} 
Identity normalization aims to normalize the expressions of all single-frame faces onto a fixed identity, pose, and background, which is distinct from popular expression reenactment. 
Currently, most expression reenactment methods \cite{ref85,ref16,ref86,ref107,ref108,ref109} cannot achieve identity normalization because they rely on consecutive frames, which cannot be applied to single-frame datasets. 
Additionally, while a few methods \cite{ref37,ref73, ref17,ref75, ref76} can achieve identity normalization, the generated quality is often limited, making it challenging to meet the demand for high-quality expression consistency in identity normalization.

\begin{figure*}[t] 
\centering 
\includegraphics[width=0.85\textwidth]{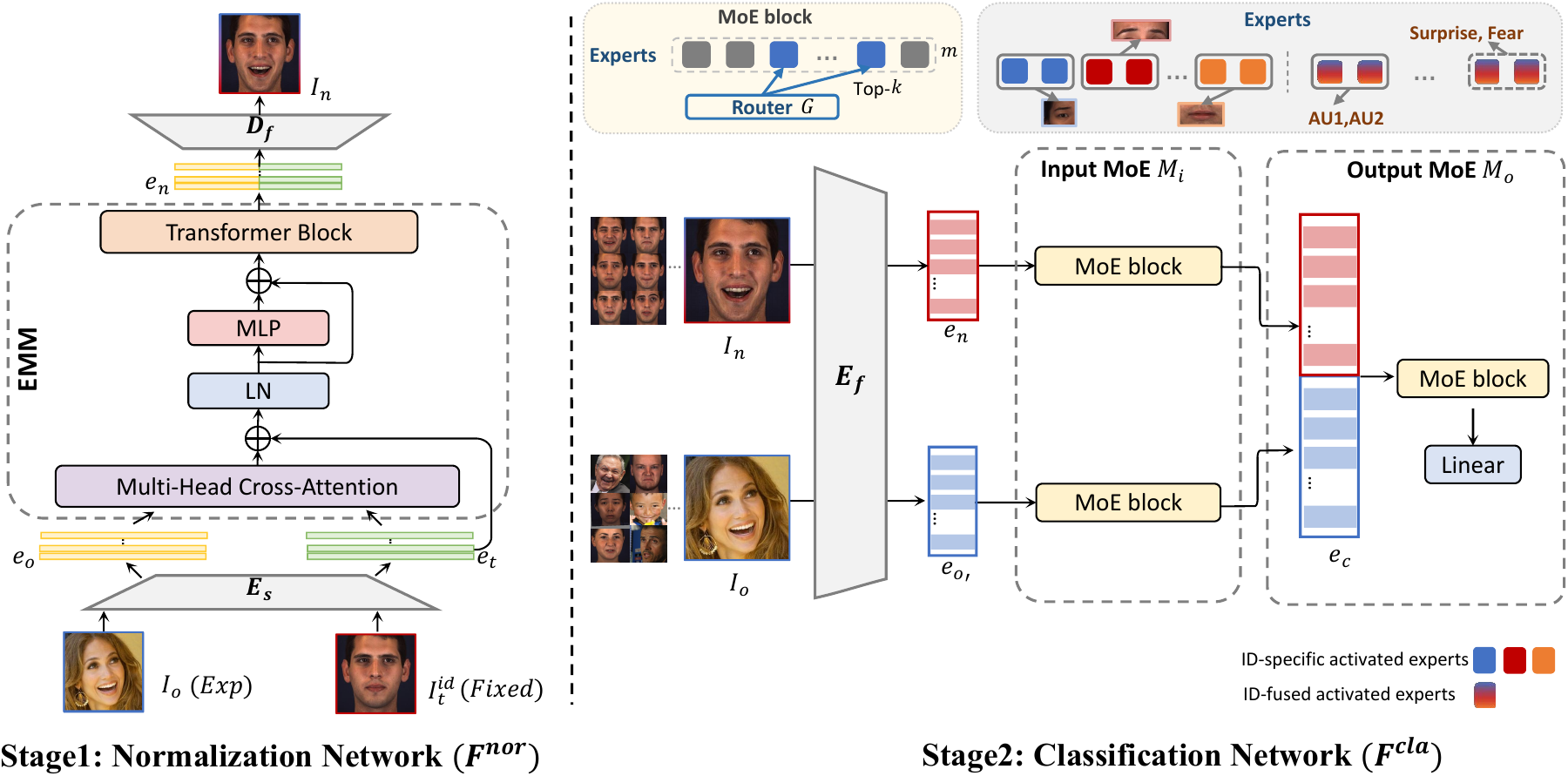}
\caption{Structure of the proposed Norface. Norface comprises two key stages: identity normalization and expression classification. In the first stage, the normalization network $F^{nor}$ normalizes all original faces to a common identity with a fixed pose and background, thereby reducing the influence of task-irrelevant factors. In the second stage, a Mixture of Expert based classification network $F^{cla}$ receives both normalized and original faces and performs AU detection, AU intensity estimation, or FER tasks.} 
\label{fig_pipeline} 
\end{figure*} 

\section{Method}
As shown in Fig.\ref{fig_pipeline}, Norface comprises two stages: the first stage employs a normalization network $F^{nor}$ for Identity normalization (Idn), while the second stage utilizes a classification network $F^{cla}$ for expression classification. Let $I_o$ be the original face belonging to a given AU or emotion dataset, and $I_{t}^{id}$ be the given fixed target face $I_{t}$ with identity $id$. Firstly, $F^{nor}$ performs Idn on $I_o$ and $I_{t}^{id}$, resulting in a normalized face $I_{n}$ that maintains consistency in terms of identity, pose, background with $I_{t}^{id}$, and expression with $I_o$. 
Then, $F^{cla}$ takes both $I_{n}$ and $I_o$ as input and then performs FEA tasks. 

\subsection{Normalization network}
Normalization network $F^{nor}$ performs Idn on the original face $I_o$ and the target face $I_t$. As shown in Fig.\ref{fig_pipeline}, a shared face encoder $E_s$ first maps $I_o$ and $I_t$ into patch embeddings $e_o$ and $e_t$. Then, an Expression Merging Module (EMM) is designed to adaptively fuse the expression information of $I_o$ and other attribute information of $I_t$. Finally, a facial decoder $D_f$ outputs the normalized face $I_n$. 

\textbf{Shared Face Encoder.} We design a shared face encoder following MAE \cite{ref38}, to project original and target faces to a common latent representation. The MAE is pre-trained on a large-scale face dataset using a mask training strategy, which has been validated in many domains due to its superior feature extraction performance \cite{ref125,ref126}. By forcing the model to learn the topological and semantic information of masked image patches, the latent space of the MAE can better capture facial expressions and attributes. 
That is, based on the pre-trained face encoder $E_s$, we can project a original/target face $I_{o/t}$ into a latent presentation:
\begin{equation}
    e_{o/t}=E_s(I_{o/t})
    \label{ep1}
\end{equation}
where $e_{o/t}\in{R^{N\times L}}$. $N$ and $L$ denote the number of patches and the dimension of each embedding, respectively.

\textbf{Expression Merging Module.} 
We designed an Expression Merging Module (EMM) that adaptively integrates the expression information of the original face with the other attributes of the target face. As shown in Fig.\ref{fig_pipeline}, EMM consists of a multi-head cross-attention block and two transformer blocks \cite{ref39}. Given the original patch embeddings $e_o$ and the target patch embeddings $e_t$, we first compute $Q$, $K$, $V$ for each patch embedding in $e_o$ and $e_t$. Then the cross-attention can be formulated as $\text{CA}\left( {Q_{t},K_{o}} \right) = \text{softmax}\left( \frac{Q_{t}{K_{o}}^{T}}{\sqrt{d_{k}}} \right)$,
where $\text{CA}$ represents Cross Attention, $Q_{*}$, $K_{*}$, $V_{*}$ are predicted by attention heads, and $d_k$ is the dimension of $K_{*}$. 
Next, the expression information from the original face is aggregated based on the computed $\text{CA}$:
\begin{equation}
    V_{fu} = \text{CA}*V_{o} + V_{t}
    \label{ep3}
\end{equation}
Then, $V_{fu}$ are normalized by a layer normalization (LN) and processed by multi-layer perceptrons (MLP). The fused embeddings $e_{fu}$ are further fed into two transformer blocks to obtain the output $e_n$. Finally, we utilize a convolutional decoder $D_f$ to generate the normalized face $I_n$ from $e_n$.

\noindent\textbf{Training Loss.} The use of multiple loss functions is a pervasive way to constrain the image generation, e.g. those works  \cite{ref122,ref123,ref127,ref130,ref131,ref132,ref133,ref134} on face manipulation, as these loss functions carry out supervision from perspectives. Inspired by them, we also employ multiple loss functions to ensure high expression consistency and fidelity in identity normalization. Specifically, the adversarial loss, the reconstruction loss, and the perceptual loss supervise the holistic image; an identity loss and a landmark loss are used to dominate the identity information; an expression loss and an eyebrow loss guide the expression transferring. They are formulated below.

\begin{equation}
\begin{aligned}
\mathcal{L}^{nor} & = \mathcal{L}_{adv} + \lambda_{rec}\mathcal{L}_{rec} + \lambda_{perc}\mathcal{L}_{perc} \\
&  + \lambda_{id}\mathcal{L}_{id} + \lambda_{lm}\mathcal{L}_{lm} + \lambda_{exp}\mathcal{L}_{exp} + \lambda_{eye}\mathcal{L}_{eye}
    \label{ep4}
\end{aligned}
\end{equation}
where $\lambda_{rec}$, $\lambda_{perc}$, $\lambda_{id}$, $\lambda_{lm}$, $\lambda_{exp}$, $\lambda_{eye}$ are hyperparameters for each term. During the training phase, the choice of $I_t$ is as random as $I_o$, which forces the normalization network to adapt to the variations in facial attributes.

\textit{Adversarial Loss.} The adversarial loss is used to make the normalized images more realistic. We apply the hinge version adversarial loss \cite{ref40} for training: 
\begin{equation}
\mathcal{L}_{adv} = - \mathbb{E}\left\lbrack {D^{res}\left( \left\lbrack {I_{t},I_{n}} \right\rbrack \right)} \right\rbrack
\end{equation}
where $D^{res}$ is the discriminator to distinguish between real and fake samples.

\textit{Reconstruction Loss.}  Since  there is no ground-truth for face normalization results, we force $I_o = I_t$ with a certain probability during training, and introduce a pixel-level reconstruction loss: 
\begin{equation}
\setlength{\abovedisplayskip}{3pt}
\setlength{\belowdisplayskip}{3pt}
\mathcal{L}_{rec} = \| I_{n}-  I_{t} \|_{2}
\end{equation}

\textit{Perceptual Loss.} Since high-level feature maps contain semantic information, we employ the feature maps from the last two convolutional layers of pre-trained VGG \cite{ref105} as the facial attribute representation. The loss is formulated as:

\begin{equation}
\mathcal{L}_{perc} = \| VGG(I_{t})  - VGG(I_{n})\|_{2}
\end{equation}

\textit{Identity Loss.} The identity loss is to constrain $I_n$’s identity information to be consistent with $I_t$:
\begin{equation}
\mathcal{L}_{id} = 1 - cos\left( E_{id}\left( I_{t} \right),E_{id}\left( I_{n} \right) \right)
\end{equation}
where $E_{id}$ denotes a face recognition model \cite{ref43} and $cos$ denotes the cosine similarity.


\textit{Landmark Loss.} To enhance facial contour consistency of $I_t$ and $I_n$, we first use the pre-trained facial landmark detector \cite{ref45} to predict the facial landmarks of $I_t$ and $I_n$, and then only apply loss to the facial contour, as follows:
\begin{equation}
\mathcal{L}_{lm} = \| P_{t}  -  P_{n}\|_{2}
\end{equation}
where $P_{t}$ and $P_{n}$ are the embeddings of the facial contours of $I_t$ and $I_n$, respectively.

\textit{Expression Loss.} To ensure that $I_o$ and $I_n$ express similar movements of full face, we use a novel fine-grained expression loss \cite{ref100} to penalize the $L_2$ distance of expression embeddings of $I_o$ and $I_n$:
\begin{equation}
\mathcal{L}_{exp} = \left\| {E_{exp}\left( I_{o} \right) - E_{exp}\left( I_{n} \right)} \right\|_{2}
\end{equation}
where $\|*\|$ denotes the euclidean distance.

\textit{Eyebrow Loss.} Since the network tends to be insensitive to subtle changes in facial eyebrows, to further supervise the expression consistency, we use expression blend shapes from 3DMM \cite{ref42}, 
and penalize the $L_2$ distance in the channel associated with eyebrow movements:
\begin{equation}
\mathcal{L}_{eye} = \left\| {E_{eye}\left( I_{o} \right) - E_{eye}\left( I_{n} \right)} \right\|_{2}
\end{equation}

\begin{figure*}[t] 
\centering 
\includegraphics[width=1\textwidth]{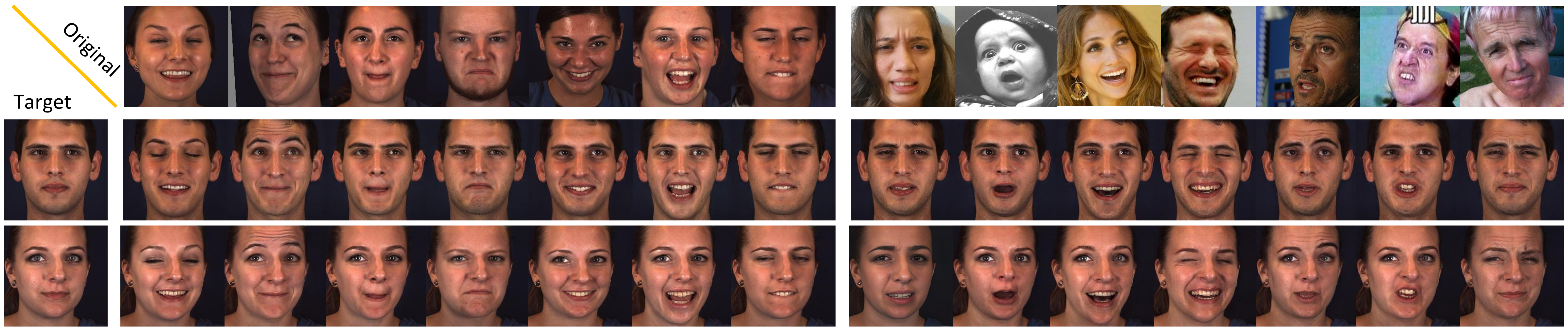}
\caption{Identity normalization of AU and emotion datasets as original faces. The figure presents results on target faces with two identities.} 
\label{fig_example} 
\end{figure*}

\noindent\textbf{Inference.} 
During inference, we set $I_o$ from a given AU or emotion dataset and provide a fixed target face $I_{t}^{id}$. As a result, we obtain normalized face $I_{n}$ that maintains consistent facial expressions as $I_o$ and ensures consistency with the identity, pose, and background of $I_{t}^{id}$. Fig.\ref{fig_example} intuitively showcases Idn results on some in-the-lab and in-the-wild AU and emotion data, and all normalized data\footnote{\href{https://norface-fea.github.io/}{Dataset link}. https://norface-fea.github.io/.} has been released.

\subsection{Classification Network}

The classification network $F^{cla}$ is designed to perform AU detection, AU intensity estimation, or emotion recognition. As shown in Fig.\ref{fig_pipeline}, $F^{cla}$ takes $I_{n}$ as input, as well as $I_o$.  
Specifically, $F^{cla}$ consists of a facial feature extractor $E_f$, an Input MoE module $M_{i}$, and an Output MoE module $M_{o}$. First, $E_f$ maps $I_{{n}}$ and $I_o$ to their latent presentations $e_{f}$, where $e_{f} \in \{e_{n},e_{o_{'}}\}$. Then, $e_{f}$ are fed into $M_{i}$, which consists of two identity-specific Mixture of Expert (MoE) \cite{ref111} blocks, resulting in the identity-specific facial representation $e_{h}$. Finally, all of $e_{h}$ are input into $M_{o}$, which refines an identity-fused representation from fixed and original identities for the final classification. $E_f$, $M_{i}$, and $M_{o}$ will be further described below, as well as the MoE block. 

${E_f}$ aims to map ${I_{n}}$ and $I_o$ to a latent representation $e_{f}$. The design of $E_f$ is similar to ${E_s}$ and it is also pre-trained on the same large-scale face datasets. That is, based on ${E_f}$, we can project $I_{{n/o}}$ into a latent presentation $e_{f}$, that is $e_{f}=E_f(I_{{n/o}})$.

$M_i$ consists of two identity-specific MoE blocks, with each MoE block receiving the individual latent representations $e_{f}$ of ${I_{n}}$ and $I_o$, respectively, that is $e_{h}=M_i(e_{f})$.

$M_o$ combines the latent representations $e_{h}$ of all identities to induce label-related information. First, all $e_h$ are concatenated to obtain the embedding $e_c$. Then, $e_{c}$ is input into an MoE block to obtain the identity-fused latent representation. Finally, we use a linear layer to output the AU detection, AU intensity estimation, or FER result.

\noindent\textit{Mixture of Experts.}
As mentioned above, MoE is used multiple times in $M_i$ and $M_o$. The MoE divides a model into a group of experts, with each expert learning individual roles, allowing for collaboration when it contributes to the task. The sparsity of these experts enhances the model's inductive capabilities \cite{ref47,ref111}, thereby effectively refining the latent representations. Specifically, the MoE in $M_i$ focuses on refining the facial representation from identity, while the MoE in $M_o$ fuses these representations from identities and focuses on refining representations relevant to different task labels. 

In our design, MoE block learns $m$ experts $E_{xp}\in\{E_{{xp}_1}, E_{{xp}_2},..., E_{{xp}_m}\}$ and a routing network $G$, where each expert is an MLP layer and multiple experts consist into a group of parallel MLP layers. By training, each expert plays individual roles but collaboratively performs their common task. In inference, $G$ assigns weights for each expert according to the MoE input.

Formally, for given input presentation $x$, the noisy \cite{ref47} Top-$k$ routing network $G$, with parameters $W_g$ and $W_{noise}$, predicts the gating weights for $x$ and selects the Top-$k$ experts from the $m$ available experts to contribute to the final representation, that is $G(x) = {\rm Topk}({\rm Softmax}(xW_{g}+ \mathcal{N}(0,1){\rm Softplus}(xW_{noise})))$.

The final representation $y$ of the $k$ activated experts are linearly combined according to the gating weights, that is $ y = {\sum\limits_{k = 1}^{m}{G^{k}(x)E_{{xp}_{k}}(x)}}$.

\noindent\textbf{Training Loss.} 
To implement expression analysis tasks, we employ four loss functions to train our $F^{cla}$:
\begin{equation}
\mathcal{L}^{cla} = \mathcal{L}_{cla} + {\lambda_{imp}\mathcal{L}}_{imp} + \lambda_{g\& l}\left( \mathcal{L}_{global} + \mathcal{L}_{local} \right)
\end{equation}
where $\lambda_{imp}$ and $\lambda_{g\& l}$ are hyperparameters for each term. $\mathcal{L}_{cla}$ is the cross entropy loss for AU detection and emotion recognition, or the L2 loss for AU intensity estimation. The importance loss $\mathcal{L}_{imp}$ is a classical solution \cite{ref47,ref48} in MoE, used to prevent the collapse caused by the Top-$k$ strategy. The global \& local loss $\mathcal{L}_{global}$ and $\mathcal{L}_{local}$  are employed to enhance the optimal utilization of experts \cite{ref49}.


\textit{Importance Loss.} The importance loss $\mathcal{L}_{imp}$, promotes a balanced distribution of the gating weights among the experts. More formally, for a given input $x$ in the batch $B$, the loss $\mathcal{L}_{imp}$ is defined by the squared coefficient of variation:
\begin{equation}
\mathcal{L}_{imp} =( \frac{{\rm std}( G(x) )}{{\rm mean}( G(x) )} )^{2}
\end{equation}

\textit{Global \& local Loss.}
To utilize each expert optimally, the global loss $\mathcal{L}_{global}$, maximizes the marginal entropy, aiming for wide expert use from a global perspective.  On the other hand, the local loss $\mathcal{L}_{local}$, encourages low entropy in focused routing weights, fostering specialization of each expert in specific tasks from a local perspective.
Formally, $p(E_{xp}|x)\in\mathbb{R}^E$ is the probability distribution of the router over $E_{xp}$ experts, $\mathcal{L}_{global}$ and $\mathcal{L}_{local}$ are defined as:
\begin{equation}
\setlength{\abovedisplayskip}{3pt}
\setlength{\belowdisplayskip}{3pt}
\begin{aligned}
\mathcal{L}_{global} = - \mathcal{H}\left( \overset{\sim}{p}\left( E_{xp} \middle| x \right) \right), 
\mathcal{L}_{local} = \frac{1}{B}{\sum\limits_{i = 1}^{B}{\mathcal{H}\left( p\left( E_{xp} \middle| x_{i} \right) \right)}} 
\end{aligned}
\end{equation}
where $\overset{\sim}{p}(E_{xp}|x)=\frac{1}{B}{\sum_{i = 1}^{B}( p( E_{xp} | x_{i}))}$ is the expert probability distribution averaged over the input $x$, and $\mathcal{H}(p)$ denotes the entropy.

\section{Experiment}

\noindent\textbf{Datasets.}
Previous works on collecting datasets (e.g., BP4D \cite{ref50}, BP4D+ \cite{ref51}, DISFA \cite{ref113}, AffectNet \cite{ref53} and RAF-DB \cite{ref52}) make significant contributions to academic communities.
We evaluate the AU analysis task on BP4D, BP4D+, DISFA and the FER task on AffectNet and RAF-DB. 
\textbf{BP4D} consists of 328 video clips from 41 participants.
About 140,000 frames are annotated with the occurrence or absence of 12 AUs for AU detection and the intensity of 5 AUs for AU intensity estimation.
\textbf{BP4D+} contains 1,400 video clips from 140 participants. 
About 198,000 frames are annotated with the same AUs in BP4D for AU detection.
\textbf{DISFA} contains about 131,000 frames from 27 video clips. Each frame of DISFA is annotated with the occurrence or absence of 8 AUs for AU detection and the intensity of 12 AUs for AU intensity estimation.
\textbf{AffectNet}, we use 7 basic emotions, selecting approximately 280,000 and 3,500 images in total for training and testing, respectively.
\textbf{RAF-DB}, the experiment's training set and test set sizes are 12,271 and 3,068, respectively. 

For fair comparisons, our experiments use the same training and testing datasets, as well as evaluation criteria as other methods. In the \textit{AU detection} task, our evaluation follows the prior works \cite{ref18,ref19}, using three-fold cross-validation, and the report results are the F1 averages from three-fold experiments.
For \textit{AU intensity estimation} task, we evaluate the model performance with intra-class correlation (ICC) \cite{ref114}, mean squared error (MSE), and mean absolute error (MAE). We report the accuracy in the \textit{FER} task.

To pick up the appropriate target face, we intentionally select 10 individuals from BP4D\&BP4D+, ensuring diversity in age, gender, and ethnicity, and used the individual with the best AU detection performance as the fixed target face $I_{t}^{id}$ that can be seen in Fig.\ref{fig_intro}. 

\noindent\textbf{Implementation Details.} 
The normalization network $F^{nor}$ is first pre-trained on CelebA-WebFace \cite{ref58} and Emo135 \cite{ref55} datasets. The face images are aligned and cropped to a size of 256$\times$256. 
Afterward, we fine-tune the network on the target AU and emotion datasets. We adopt Adam \cite{ref112} optimizer with $\beta_1=0$, $\beta_2=0.99$, a learning rate of 0.0001, and a batch size of 8. We set $\lambda_{rec}=10$, $\lambda_{perc}=5$, $\lambda_{id}=10$, $\lambda_{lm}=5000$, $\lambda_{exp}=5000$, and $\lambda_{eye}=10$.
For the classification network $F^{cla}$, 
we performed fine-tuning for 40 epochs, and set $\lambda_{imp}=0.001$, $\lambda_{g\& l}=0.001$. The settings for each MoE block are $m=4$ and $k=2$. The base learning rates are set to $1e-4$, $1e-3$, and $2e-5$ for AU detection, AU intensity estimation, and FER tasks, respectively.

\begin{table*}[t] 
\setlength{\belowcaptionskip}{0.2cm}
\tiny
\setlength{\arrayrulewidth}{0.3pt}
\centering
\begin{tablefont}
\setlength{\tabcolsep}{0.9mm}{
\caption{Comparison of AU detection results on BP4D and BP4D+ in terms of F1 scores (\%).}
\label{tab1}
\begin{tabular}{c|c|ccccccccccccc|c} 
Dataset & Method       & AU1           & AU2           & AU4           & AU6           & AU7           & AU10          & AU12          & AU14          & AU15          & AU17          & AU23          & AU24          & Avg.      & $\Delta\uparrow$    \\ 
\specialrule{0.6pt}{0pt}{0pt}
\multirow{9}{*}{BP4D} &
LP-Net \cite{ref5}      & 43.4          & 38.0          & 54.2          & 77.1          & 76.7          & 83.8          & 87.2          & 63.3          & 45.3          & 60.5          & 48.1          & 54.2          & 61.0    & 0.0       \\
  & JAA-Net \cite{ref19}     & 53.8          & 47.8          & 58.2          & 78.5          & 75.8          & 82.7          & 88.2          & 63.7          & 43.3          & 61.8          & 45.6          & 49.9          & 62.4    & +1.4       \\
& HMP-PS \cite{ref59}      & 53.1          & 46.1          & 56.0          & 76.5          & 76.9          & 82.1          & 86.4          & 64.8          & 51.5          & 63.0          & [49.9]          & 54.5          & 63.4     & +2.4      \\
& SEV-Net \cite{ref7}     & 58.2      & [50.4]      & 58.3          & \textbf{81.9} & 73.9          & \textbf{87.8} & 87.5          & 61.6          & 52.6          & 62.2          & 44.6          & 47.6          & 63.9       & +2.9    \\
& MHSA-FFN \cite{ref2}    & 51.7          & 49.3          & [61.0]          & 77.8          & 79.5          & 82.9          & 86.3          & {[}67.6]          & 51.9          & 63.0          & 43.7          & 56.3          & 64.2       & +3.2    \\
& CaFNet \cite{ref60}      & 55.1          & 49.3          & 57.7          & 78.3          & 78.6          & 85.1          & 86.2          & 67.4          & 52.0          & 64.4      & 48.3          & 56.2          & 64.9       & +3.9    \\
& Chang et al. \cite{ref61} & 53.3          & 47.4          & 56.2          & 79.4          & \textbf{80.7} & 85.1          & {[}89.0]          & 67.4          & {[}55.9]          & 61.9          & 48.5          & 49.0          & 64.5        & +3.5   \\
& GLEE-Net \cite{ref62}  & {[}60.6]          & 44.4          & 61.0        & {[}80.6]      & 78.7      & 85.4          & 88.1      & 64.9      & 53.7      & {[}65.1]      & 47.7          & {[}58.5]      & [65.7]     & +4.7  \\ 
&{\cellcolor[rgb]{0.902,0.902,0.902}} Norface (Ours)        &{\cellcolor[rgb]{0.902,0.902,0.902}} \textbf{60.9} &{\cellcolor[rgb]{0.902,0.902,0.902}} \textbf{55.3} &{\cellcolor[rgb]{0.902,0.902,0.902}} \textbf{67.4} &{\cellcolor[rgb]{0.902,0.902,0.902}} 79.7      &{\cellcolor[rgb]{0.902,0.902,0.902}} {[}79.9]          &{\cellcolor[rgb]{0.902,0.902,0.902}} {[}87.5]      &{\cellcolor[rgb]{0.902,0.902,0.902}} \textbf{90.7} &{\cellcolor[rgb]{0.902,0.902,0.902}} \textbf{69.8} &{\cellcolor[rgb]{0.902,0.902,0.902}} \textbf{58.1} &{\cellcolor[rgb]{0.902,0.902,0.902}} \textbf{66.2} &{\cellcolor[rgb]{0.902,0.902,0.902}} \textbf{54.9} &{\cellcolor[rgb]{0.902,0.902,0.902}} \textbf{61.8} &{\cellcolor[rgb]{0.902,0.902,0.902}} \textbf{69.3} &{\cellcolor[rgb]{0.902,0.902,0.902}}\textbf{+8.3} \\
\specialrule{0.3pt}{0pt}{0pt}
\multirow{5}{*}{BP4D+} & ML-GCN \cite{ref63}   & 40.2          & 36.9          & 32.5          & 84.8          & 88.9      & 89.6          & 89.3          & 81.2          & {[}53.3]      & 43.1          & 55.9          & 28.3          & 60.3      & 0.0     \\
& MS-CAM \cite{ref64}  & 38.3          & 37.6          & 25.2          & 85.0          & \textbf{90.9} & {[}90.9]      & 89.0          & 81.5          & \textbf{60.9} & 40.6          & 58.2          & 28.0          & 60.5         & +0.2  \\
& SEV-Net \cite{ref7}  & 47.9          & 40.8          & 31.2          & {[}86.9]      & 87.5          & 89.7          & 88.9          & {[}82.6]      & 39.9          & \textbf{55.6} & [59.4]      & 27.1          & 61.5        & +1.2   \\
& GLEE-Net \cite{ref62}  & \textbf{54.2} & \textbf{46.3} & [38.1]      & 86.2          & 87.6          & 90.4          & {[}89.5]      & 81.3          & 46.3          & 47.4          & 57.6          & [39.6]      & [63.7]     & +3.4  \\ 
&{\cellcolor[rgb]{0.902,0.902,0.902}} Norface (Ours)     &{\cellcolor[rgb]{0.902,0.902,0.902}} {[}52.2]      &{\cellcolor[rgb]{0.902,0.902,0.902}} {[}46.0]      &{\cellcolor[rgb]{0.902,0.902,0.902}} \textbf{51.7} &{\cellcolor[rgb]{0.902,0.902,0.902}} \textbf{88.1} &{\cellcolor[rgb]{0.902,0.902,0.902}} {[}89.0]          &{\cellcolor[rgb]{0.902,0.902,0.902}} \textbf{91.3} &{\cellcolor[rgb]{0.902,0.902,0.902}} \textbf{90.1} &{\cellcolor[rgb]{0.902,0.902,0.902}} \textbf{83.3} &{\cellcolor[rgb]{0.902,0.902,0.902}} 50.4          &{\cellcolor[rgb]{0.902,0.902,0.902}} {[}51.1]      &{\cellcolor[rgb]{0.902,0.902,0.902}} \textbf{61.6} &{\cellcolor[rgb]{0.902,0.902,0.902}} \textbf{45.4} &{\cellcolor[rgb]{0.902,0.902,0.902}} \textbf{66.7}& {\cellcolor[rgb]{0.902,0.902,0.902}}\textbf{+6.4}\\
\end{tabular}}
\end{tablefont}
\end{table*}

\begin{table}
\centering
\tiny
\setlength{\abovecaptionskip}{0cm} 
\setlength{\belowcaptionskip}{0.2cm}
\begin{tablefont}
\setlength{\tabcolsep}{2.0mm}{
\caption{Comparison of AU detection results on DISFA in terms of F1 scores (\%).}
\label{tabdisfa}
\begin{tabular}{ccccccccccc}
Method   & AU1  & AU2  & AU4  & AU6  & AU9  & AU12 & AU25 & AU26  & Avg. &{\cellcolor[rgb]{0.902,0.902,0.902}} $\Delta\uparrow$   \\ 
\specialrule{0.6pt}{0pt}{0pt}
LP-Net \cite{ref5}   & 29.9 & 24.7 & 72.7 & [46.8] & 49.6 & 72.9 & 93.8 & 56.0  & 56.9 &{\cellcolor[rgb]{0.902,0.902,0.902}} 0.0  \\
JAA-Net \cite{ref19}  & 62.4 & 60.7 & 67.1 & 41.1 & 45.1 & 73.5 & 90.9 & 67.4  & 63.5 &{\cellcolor[rgb]{0.902,0.902,0.902}} +6.6  \\
GLEE-Net \cite{ref62} & 61.9 & 54.0 & \textbf{75.8} & 45.9 & [55.7] & [77.6] & 92.9 & 60.0  & 65.5 &{\cellcolor[rgb]{0.902,0.902,0.902}} +8.6  \\
LGRNet \cite{ref115}  & [62.6] & [64.4] & 72.5 & 46.6 & 48.8 & 75.7 & [94.4] & \textbf{73.0} & [67.3] &{\cellcolor[rgb]{0.902,0.902,0.902}} +10.4  \\ 
\specialrule{0.2pt}{0pt}{0pt}
 Norface (Ours)   & \textbf{76.4}     & \textbf{66.1}     & [74.2]     & \textbf{58.5}     & \textbf{57.2}     & \textbf{81.7}     & \textbf{97.6}     & [69.6]      & \textbf{72.7} &{\cellcolor[rgb]{0.902,0.902,0.902}} \textbf{+15.8}  \\
\end{tabular}}
\end{tablefont}
\end{table}

\subsection{Comparison with State-of-the-Art Methods}
We verify the effectiveness of our method on AU detection, AU intensity estimation, and facial emotion recognition tasks, as well as their cross-dataset tasks.

\noindent\textbf{AU detection task.} 
Tab.\ref{tab1} and \ref{tabdisfa} report comparison results with those previous methods. Our method outperforms those baselines on all datasets with the average of AUs detection. Specifically, ours exceeds the SOTA of GLEE-Net \cite{ref62} for over 3\% on both BP4D and BP4D+ datasets, and exceeds the SOTA of LGRNet \cite{ref115} for over 5\% on the DISFA dataset, which validates our method on the AU detection task.

\noindent\textbf{AU intensity estimation.} 
Tab.\ref{tabins} report comparison results with those previous methods. Ours surpasses all existing approaches in terms of MSE and MAE, and exceeds the SOTA of APs \cite{ref119} by 0.19 in terms of ICC on the DISFA dataset, which validates our method on the AU intensity estimation task.

\begin{table}[t] 
\centering
\tiny
\setlength{\arrayrulewidth}{0.3pt}
\setlength{\belowcaptionskip}{0.2cm}
\begin{tablefont}
\setlength{\tabcolsep}{0.7mm}{
\caption{Comparison of AU  intensity estimation results on BP4D and DISFA.}
\label{tabins}
\begin{tabular}{c|c|cccccc|ccccccccccccc} 
\multirow{2}{*}{Metric} & \multirow{2}{*}{Method} & \multicolumn{6}{c|}{AU on BP4D}             & \multicolumn{13}{c}{AU on DISFA}                                               \\
                        &                         & 6   & 10  & 12  & 14   & 17  & Avg.         & 1   & 2   & 4    & 5   & 6   & 9   & 12  & 15  & 17  & 20  & 25  & 26  & Avg.  \\ 
\specialrule{0.6pt}{0pt}{0pt}
\multirow{5}{*}{ICC $\uparrow$}    & ISIR \cite{ref116}                    & .79 & .80 & .86 & .71  & .44 & .72          & -   & -   & -    & -   & -   & -   & -   & -   & -   & -   & -   & -   & -     \\
                        & HR  \cite{ref117}                    & .82 & .80 & .86 & .69  & .51 & .73          & .35 & .19 & .78  & .73 & .52 & .65 & .81 & .49 & .61 & .28 & .92 & .67 & .58   \\
                        & SCC-Heatmap  \cite{ref118}          & .74 & .82 & .86 & .68  & .51 & .72          & .73 & .44 & .74  & .06 & .27 & .51 & .71 & .04 & .37 & .04 & .94 & .78 & .47   \\
                        & APs \cite{ref119}                    & .82 & .80 & .86 & .69  & .51 & \textbf{.74} & .35 & .19 & .78  & .73 & .52 & .65 & .81 & .49 & .61 & .28 & .92 & .67 & .58   \\
                        &{\cellcolor[rgb]{0.902,0.902,0.902}} Norface (Ours)                 &{\cellcolor[rgb]{0.902,0.902,0.902}} .81 &{\cellcolor[rgb]{0.902,0.902,0.902}} .74 &{\cellcolor[rgb]{0.902,0.902,0.902}} .90 &{\cellcolor[rgb]{0.902,0.902,0.902}} .50  &{\cellcolor[rgb]{0.902,0.902,0.902}} .74 &{\cellcolor[rgb]{0.902,0.902,0.902}} \textbf{.74} &{\cellcolor[rgb]{0.902,0.902,0.902}} .72    &{\cellcolor[rgb]{0.902,0.902,0.902}} .68    &{\cellcolor[rgb]{0.902,0.902,0.902}} .77     &{\cellcolor[rgb]{0.902,0.902,0.902}} .68    &{\cellcolor[rgb]{0.902,0.902,0.902}} .59    &{\cellcolor[rgb]{0.902,0.902,0.902}} .56    &{\cellcolor[rgb]{0.902,0.902,0.902}} .87    &{\cellcolor[rgb]{0.902,0.902,0.902}} .54    &{\cellcolor[rgb]{0.902,0.902,0.902}} .65    &{\cellcolor[rgb]{0.902,0.902,0.902}} .34    &{\cellcolor[rgb]{0.902,0.902,0.902}} .96    &{\cellcolor[rgb]{0.902,0.902,0.902}}  .70   &{\cellcolor[rgb]{0.902,0.902,0.902}} \textbf{.67}      \\ 
\specialrule{0.3pt}{0pt}{0pt}
\multirow{4}{*}{MSE $\downarrow$}    & ISIR \cite{ref116}                   & .83 & .80 & .62 & 1.14 & .84 & .85          & -   & -   & -    & -   & -   & -   & -   & -   & -   & -   & -   & -   & -     \\
                        & HR \cite{ref117}                     & .68 & .80 & .79 & .98  & .61 & .78          & .41 & .37 & .70  & .08 & .44 & .30 & .29 & .14 & .26 & .16 & .24 & .39 & .32   \\
                        & APs \cite{ref119}                    & .72 & .84 & .60 & 1.13 & .57 & .77          & .68 & .59 & .40  & .03 & .49 & .15 & .26 & .13 & .22 & .20 & .35 & .17 & .30   \\
                        &{\cellcolor[rgb]{0.902,0.902,0.902}} Norface (Ours)                 &{\cellcolor[rgb]{0.902,0.902,0.902}} .71 &{\cellcolor[rgb]{0.902,0.902,0.902}} .95 &{\cellcolor[rgb]{0.902,0.902,0.902}} .50 &{\cellcolor[rgb]{0.902,0.902,0.902}} 1.02 &{\cellcolor[rgb]{0.902,0.902,0.902}} .47 &{\cellcolor[rgb]{0.902,0.902,0.902}} \textbf{.73} &{\cellcolor[rgb]{0.902,0.902,0.902}} .22    &{\cellcolor[rgb]{0.902,0.902,0.902}} .19    &{\cellcolor[rgb]{0.902,0.902,0.902}} .46     &{\cellcolor[rgb]{0.902,0.902,0.902}} .03    &{\cellcolor[rgb]{0.902,0.902,0.902}} .41    &{\cellcolor[rgb]{0.902,0.902,0.902}} .30    &{\cellcolor[rgb]{0.902,0.902,0.902}} .24     &{\cellcolor[rgb]{0.902,0.902,0.902}}  .09   &{\cellcolor[rgb]{0.902,0.902,0.902}} .17    &{\cellcolor[rgb]{0.902,0.902,0.902}} .14    &{\cellcolor[rgb]{0.902,0.902,0.902}} .15    &{\cellcolor[rgb]{0.902,0.902,0.902}} .29    &{\cellcolor[rgb]{0.902,0.902,0.902}} \textbf{.22}      \\ 
\specialrule{0.3pt}{0pt}{0pt}
\multirow{4}{*}{MAE $\downarrow$}    & BORMIR \cite{ref121}                 & .85 & .90 & .68 & 1.05 & .79 & .85          & .88 & .78 & 1.24 & .59 & .77 & .78 & .76 & .56 & .72 & .63 & .90 & .88 & .79   \\
                        & KBSS \cite{ref120}                   & .65 & .65 & .48 & .98  & .63 & .66          & .48 & .49 & .57  & .08 & .26 & .22 & .33 & .15 & .44 & .22 & .43 & .36 & .33   \\
                        & SCC-Heatmap \cite{ref119}            & .61 & .56 & .52 & .73  & .50 & .58          & .16 & .16 & .27  & .03 & .25 & .13 & .32 & .15 & .20 & .09 & .30 & .32 & .20   \\
                        &{\cellcolor[rgb]{0.902,0.902,0.902}} Norface (Ours)                 &{\cellcolor[rgb]{0.902,0.902,0.902}} .48 &{\cellcolor[rgb]{0.902,0.902,0.902}} .56 &{\cellcolor[rgb]{0.902,0.902,0.902}} .39 &{\cellcolor[rgb]{0.902,0.902,0.902}} .82  &{\cellcolor[rgb]{0.902,0.902,0.902}} .40 &{\cellcolor[rgb]{0.902,0.902,0.902}} \textbf{.53} &{\cellcolor[rgb]{0.902,0.902,0.902}} .14    &{\cellcolor[rgb]{0.902,0.902,0.902}} .12    &{\cellcolor[rgb]{0.902,0.902,0.902}} .30     &{\cellcolor[rgb]{0.902,0.902,0.902}} .03    &{\cellcolor[rgb]{0.902,0.902,0.902}}  .29   &{\cellcolor[rgb]{0.902,0.902,0.902}} .17    &{\cellcolor[rgb]{0.902,0.902,0.902}} .22    &{\cellcolor[rgb]{0.902,0.902,0.902}} .09    &{\cellcolor[rgb]{0.902,0.902,0.902}} .14    &{\cellcolor[rgb]{0.902,0.902,0.902}} .08   &{\cellcolor[rgb]{0.902,0.902,0.902}} .17    &{\cellcolor[rgb]{0.902,0.902,0.902}} .26   &{\cellcolor[rgb]{0.902,0.902,0.902}}   \textbf{.17}    \\
\end{tabular}}
\end{tablefont}
\end{table}

\begin{table}[t] 
 \begin{minipage}[t]{0.42\textwidth}
\centering
\tiny
\captionof{table}{Comparison of facial emotion recognition results (\%).}
\renewcommand\arraystretch{1.1}
   \setlength{\arrayrulewidth}{0.3pt}
\begin{tablefont}
\setlength{\tabcolsep}{0.8mm}{
\label{tabfer}
\begin{tabular}{c|cc|cc} 
\multirow{2}{*}{Method}  & \multicolumn{2}{c|}{RAF-DB}                                & \multicolumn{2}{c}{AffectNet}                                   \\ 
                      & \multicolumn{1}{c}{Acc.} & \multicolumn{1}{c|}{{\cellcolor[rgb]{0.902,0.902,0.902}}$\Delta\uparrow$} & \multicolumn{1}{c}{Acc.} & \multicolumn{1}{c}{{\cellcolor[rgb]{0.902,0.902,0.902}}$\Delta\uparrow$} \\ 
\specialrule{0.6pt}{0pt}{0pt}
LDL-ALSG \cite{ref9}     & 85.53    & {\cellcolor[rgb]{0.902,0.902,0.902}}0.00     & 59.35      & {\cellcolor[rgb]{0.902,0.902,0.902}}0.00       \\
DLN \cite{ref100} & 86.40 & {\cellcolor[rgb]{0.902,0.902,0.902}}+0.87 & 63.70 &{\cellcolor[rgb]{0.902,0.902,0.902}}+4.35 \\
SCN \cite{ref25}          & 87.03    & {\cellcolor[rgb]{0.902,0.902,0.902}}+1.53     & -         & {\cellcolor[rgb]{0.902,0.902,0.902}}-         \\
PSR \cite{ref28}          & 88.98    & {\cellcolor[rgb]{0.902,0.902,0.902}}+3.45     & 63.77     & {\cellcolor[rgb]{0.902,0.902,0.902}}+4.42         \\
DMUE \cite{ref23}         & 89.42    & {\cellcolor[rgb]{0.902,0.902,0.902}}+3.89     & 63.11     & {\cellcolor[rgb]{0.902,0.902,0.902}}+3.76         \\
FDRL \cite{ref6}         & 89.47    & {\cellcolor[rgb]{0.902,0.902,0.902}}+3.94     & -         & {\cellcolor[rgb]{0.902,0.902,0.902}}-         \\
TransFER \cite{ref27}     & 90.91    & {\cellcolor[rgb]{0.902,0.902,0.902}}+5.38     & 66.23     & {\cellcolor[rgb]{0.902,0.902,0.902}}+6.88         \\
Meta-Face2Exp \cite{ref1} & 88.54    & {\cellcolor[rgb]{0.902,0.902,0.902}}+3.01     & 64.23     & {\cellcolor[rgb]{0.902,0.902,0.902}}+4.88         \\
Latent-OFER \cite{ref3}  & 89.60 & {\cellcolor[rgb]{0.902,0.902,0.902}}+4.07  & 63.90 & {\cellcolor[rgb]{0.902,0.902,0.902}}+4.55 \\
LA-Net \cite{ref8} & [91.56]    & {\cellcolor[rgb]{0.902,0.902,0.902}}+6.03     & [67.60]     & {\cellcolor[rgb]{0.902,0.902,0.902}}+8.25        \\
\specialrule{0.3pt}{0pt}{0pt}
Norface (Ours)           & \textbf{92.97} & {\cellcolor[rgb]{0.902,0.902,0.902}}\textbf{+7.44}  & \textbf{68.69}   & {\cellcolor[rgb]{0.902,0.902,0.902}}\textbf{+9.34}   \\
    \end{tabular}}
    \end{tablefont}
        \end{minipage}
        \hfill
  \begin{minipage}[t]{0.55\textwidth} 
   \centering
   \setlength{\arrayrulewidth}{0.3pt}
   \begin{tablefont}
\setlength{\abovecaptionskip}{0cm} 
\captionof{table}{Comparison of AU detection on cross-dataset (trained on BP4D and evaluated on BP4D+) in terms of F1 scores (\%).}
\tiny
\setlength{\tabcolsep}{0.5mm}{
\label{tab6}
\begin{tabular}{c|c|c|c|c} 
Method & EAC-Net \cite{ref18} & JAA-Net \cite{ref19}      & GLEE-Net \cite{ref62}     &{\cellcolor[rgb]{0.902,0.902,0.902}} Norface (Ours)        \\
\specialrule{0.6pt}{0pt}{0pt}
AU1    & 38.0     & 39.7          & {[}39.8]      &{\cellcolor[rgb]{0.902,0.902,0.902}} \textbf{48.2}  \\
AU2    & 37.5     & 35.6          & {[}37.9]      &{\cellcolor[rgb]{0.902,0.902,0.902}} \textbf{39.1}  \\
AU4    & 32.6     & 30.7          & {[}41.6]      &{\cellcolor[rgb]{0.902,0.902,0.902}} \textbf{47.1}  \\
AU6    & 82.0     & 82.4          & {[}83.4]      &{\cellcolor[rgb]{0.902,0.902,0.902}} \textbf{83.5}  \\
AU7    & 83.4     & 84.7          & {[}88.2]      &{\cellcolor[rgb]{0.902,0.902,0.902}} \textbf{89.7}  \\
AU10   & 87.1     & 88.8          & \textbf{90.2} &{\cellcolor[rgb]{0.902,0.902,0.902}} {[}89.4]       \\
AU12   & 85.1     & 87.0          & \textbf{86.9} &{\cellcolor[rgb]{0.902,0.902,0.902}} {[}86.3]       \\
AU14   & 62.1     & 62.2          & {[}76.6]      &{\cellcolor[rgb]{0.902,0.902,0.902}} \textbf{81.1}  \\
AU15   & {[}44.5] & 38.9          & \textbf{48.3} &{\cellcolor[rgb]{0.902,0.902,0.902}} 43.2           \\
AU17   & 43.6     & \textbf{46.4} & 42.9          &{\cellcolor[rgb]{0.902,0.902,0.902}} {[}45.4]       \\
AU23   & 45.0     & {[}48.9]      & 47.7          &{\cellcolor[rgb]{0.902,0.902,0.902}} \textbf{63.2}  \\
AU24   & 32.8     & \textbf{36.6} & 29.8          &{\cellcolor[rgb]{0.902,0.902,0.902}} {[}35.1]       \\
\specialrule{0.3pt}{0pt}{0pt}
Avg.   & 56.1     & 56.8 & [57.5]          &{\cellcolor[rgb]{0.902,0.902,0.902}} \textbf{62.6}       \\
      \end{tabular}}
      \end{tablefont}
   \end{minipage}
\end{table}

\noindent\textbf{Facial emotion recognition task.} 
Tab.\ref{tabfer} reports comparison results with previous methods. Ours outperforms these baselines in terms of accuracy on both datasets. Specifically, ours exceeds the SOTA of LA-Net \cite{ref8} for 1.41\% and 1.09\% on RAF-DB and AffectNet, respectively, validating our method on the FER task.

\noindent\textbf{Cross-dataset task.}
Tab.\ref{tab6} and \ref{tab_crossFER} report comparison results of AU detection and FER with previous methods on cross-dataset tasks, respectively. Ours achieves the best performance, surpassing the SOTA of GLEE-Net \cite{ref62} by 5.1\% on AU detection, and outperforming  RANDA \cite{ref67} by 6.09\% and 11.2\% on the 'R$\rightarrow$A' and 'A$\rightarrow$R' tasks of FER, respectively.
Due to the normalization of task-irrelevant noise, domain barriers between datasets are reduced, which highlights the benefits of our method for cross-domain tasks.

\begin{table}[t] 
 \begin{minipage}[t]{0.45\textwidth}
\centering
\tiny
 \renewcommand\arraystretch{1.1}
\setlength{\abovecaptionskip}{0cm}
\setlength{\arrayrulewidth}{0.3pt}
\begin{tablefont}
\setlength{\tabcolsep}{0.5mm}{
\caption{Comparison of FER on cross-dataset (\%). R$\rightarrow$A (A$\rightarrow$R) indicates trained on RAF-DB (AffectNet) and evaluated on AffectNet (RAF-DB).}
\label{tab_crossFER}
\begin{tabular}{c|c|c|c|c|c} 
Method & FDRL \cite{ref6} & RANDA \cite{ref67}  & TransFER \cite{ref27}      & CA-FER \cite{ref66}     & Norface          \\ 
\specialrule{0.6pt}{0pt}{0pt}
R$\rightarrow$A  & 41.57 & [52.34]     & 44.57 & 47.2          & \textbf{58.43}       \\
A$\rightarrow$R  & - & 62.48     & - & -          & \textbf{73.68}  
 \\
\end{tabular}}
\end{tablefont}
\end{minipage}
        \hfill
 \begin{minipage}[t]{0.5\textwidth}
\setlength{\abovecaptionskip}{0cm}
\tiny
\begin{tablefont}
\centering
\setlength{\arrayrulewidth}{0.3pt}
\setlength{\tabcolsep}{1.2mm}{
\caption{Results of Norface $w$ and $w/o$ Idn. AU det: AU detection.}
\label{tab3}
\begin{tabular}{c|ccc|cc}
\multirow{2}{*}{Method}  & \multicolumn{3}{c|}{AU det(\textit{F1})} & \multicolumn{2}{c}{FER(\textit{Acc.})} \\
        & BP4D & BP4D+ & DISFA & RAF-DB & AffectNet \\
\specialrule{0.6pt}{0pt}{0pt}
$w/o$ Idn & 67.2 & 64.8  & 70.2  & 89.24  & 67.29     \\
$w/$ Idn   & \textbf{69.3} & \textbf{66.7}  & \textbf{72.7}  & \textbf{92.97}  & \textbf{68.69}     \\
\end{tabular}}
\end{tablefont}
\end{minipage}
\end{table}

The above experiments validate our framework being unified for those tasks, as identity normalization addresses the interference of task-irrelevant noise, such as identity, pose, and background, etc. More experiments on identity normalization are described below.

\subsection{Analysis on Identity Normalization}
We address the following Idn questions: 
\textbf{A1.} Is Idn effective? 
\textbf{A2.} How does the normalization network perform?
\textbf{A3.} What is the difference between Idn and data augmentation?
\textbf{A4.} Why use normalized images instead of expression features for the classification network?

\noindent\textbf{A1. Impact of Idn.}
Tab.\ref{tab3} reports the results of different FEA tasks $w/$ and $w/o$ Idn. The results show that the performance $w/$ Idn consistently outperforms that $w/o$ Idn in all tasks. For example, $w/$ Idn, there are improvements of 2.1\%, 1.9\%, and 2.5\% in AU detection on BP4D, BP4D+, and DISFA, 
3.73\% and 1.4\% increase in FER on RAF-DB and AffectNet, highlighting the benefits of Idn.

\begin{figure}[t]
\centering
\begin{minipage}[t]{.48\textwidth}
\centering
\includegraphics[scale=0.22]{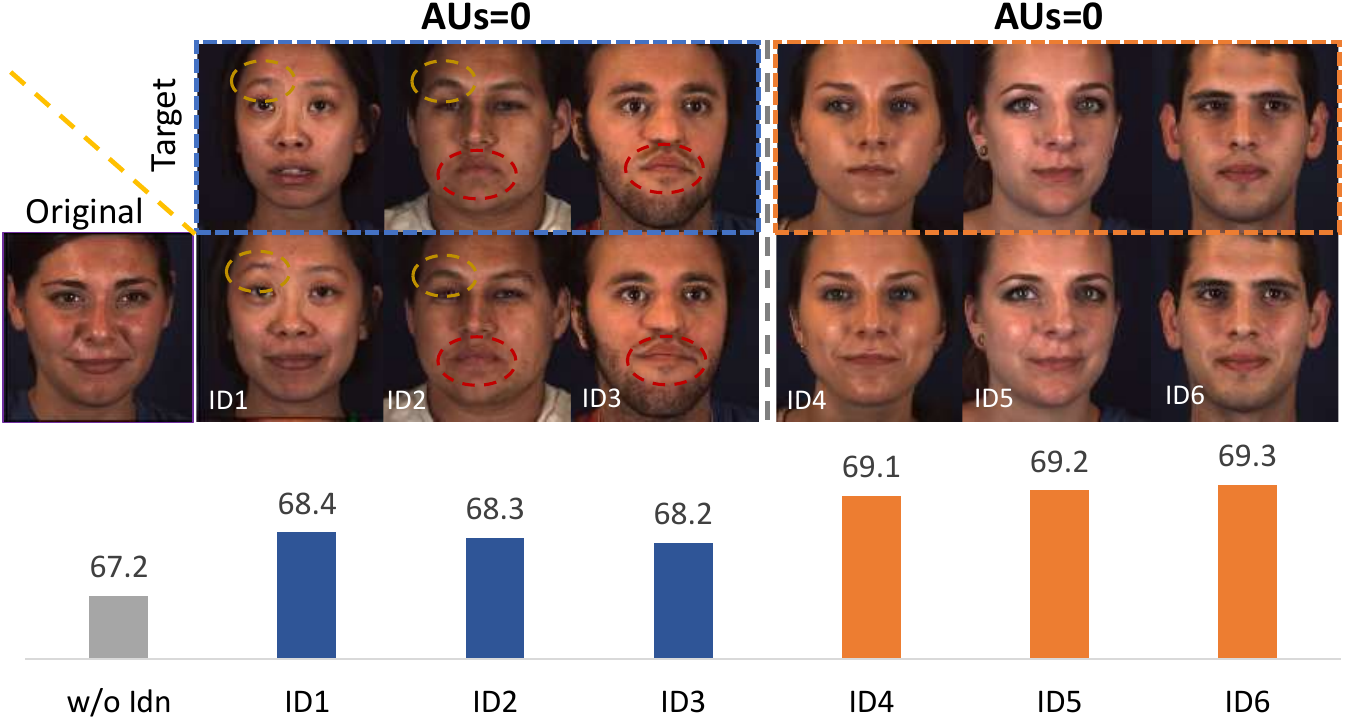}
\caption{Normalization results for different identities as target face (top) and AU detection results based on BP4D (bottom).}
\label{fig_case}
\end{minipage}
\hfill
\begin{minipage}[t]{0.48\textwidth}
\centering
\includegraphics[scale=0.22]{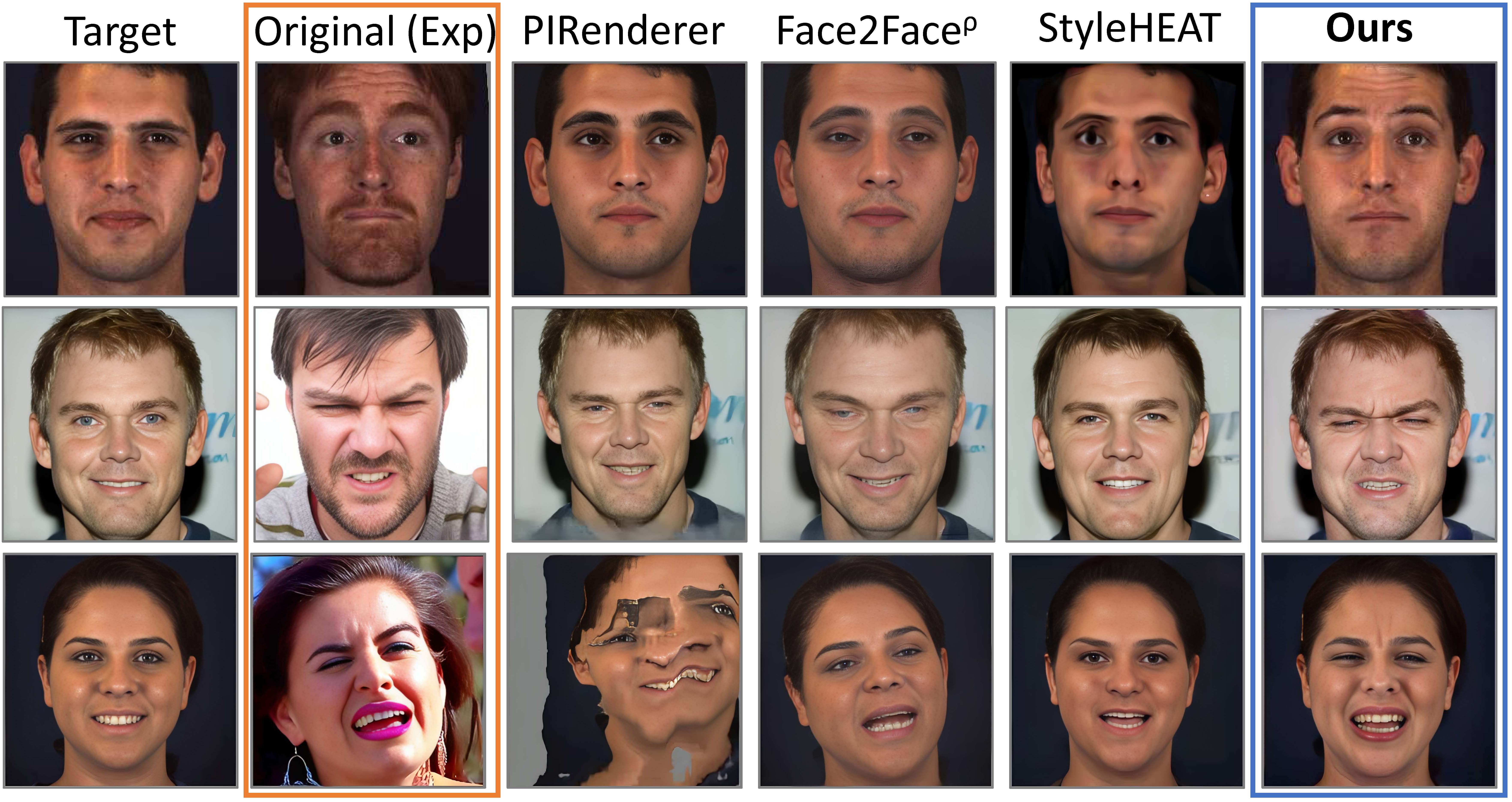}
\caption{Visualization of different methods for normalization.}
\label{fig_other}
\end{minipage}
\end{figure}

In addition, Fig.\ref{fig_case} provides a detailed presentation of the normalization and AU detection results for different target identities. It is clear that using several target identities are validated to consistently yields better results than $w/o$ Idn.
Moreover, our study indicates that the selection of target identities contributes differently to performance enhancement, and certain identities are unsuitable as target faces. For example, $w/$ 'ID3' resulted in a performance decrease of 1.1\% compared to $w/$ 'ID6', which we attribute to the \textit{confusion} in facial expressions. In Fig.\ref{fig_case}, the non-candidate target faces in the blue box are labeled as having no AUs present, but these faces still exhibit raised eyebrows or lip corner depressors, leading to confusion in the expressions of their normalized faces. In contrast, as shown in the orange box in Fig.\ref{fig_case}, these faces have minimal confusion in facial expressions, ensuring high discriminability and performance enhancement. They can be considered as candidate target faces.

Our method is compared to multi-task methods, based on BP4D and AffectNet for fair comparison, ours exceeds JPML \cite{ref13} for 23.4\% and MT-VGG \cite{ref88} for 8.69\% under BP4D and AffectNet, respectively. Compared to identity-invariant methods, ours exceeds the SOTA of DLN \cite{ref100} for 4.99\% under AffectNet and exceeds the SOTA of IPD-FER \cite{ref92} for 4.16\% under RAF-DB. This clearly illustrates the significant benefits of Idn for AU analysis and FER tasks.

\noindent\textbf{A2. Normalization performance.} Fig.\ref{fig_other} visualizes the results of our $F^{nor}$ and other available methods enabling identity normalization. Our method offers better expression consistency between output and original faces than PIRenderer \cite{ref73}, Face2Face$^\rho$ \cite{ref17}, and StyleHEAT \cite{ref75}. Additionally, a subjective evaluation with 100 volunteers rating 50 randomly selected normalized images on a scale of 0 to 5 shows our method (4.6) significantly outperforms PIRenderer (2.2), Face2Face$^\rho$ (1.8), and StyleHEAT (1.6). 
The conducted Paired Samples T-Test shows significant differences in expression consistency between ours (M = 4.6, SD = 0.49) and PIRenderer (M = 2.2, SD = 0.61), t(49) = 21.00, p$<$ 0.0001; between ours and Face2Face$^\rho$ (M = 1.8, SD = 0.61), t(49) = 26.19, p$<$ 0.0001; between ours and StyleHEAT (M = 1.6, SD = 0.49), t(49) = 33.20, p$<$ 0.0001.


\noindent\textbf{A3. Data augmentation.} Tab.\ref{tab7} shows results of recent data augmentation methods such as mixup \cite{ref68}, cutout \cite{ref72}, and cutmix \cite{ref69}. Ours outperforms these on both AU detection and FER tasks, e.g., outperforming mixup by 2.1\%, 1.9\%, 1.9\%, 3.73\%, and 1.4\% on the five datasets, respectively. 
Unlike traditional data augmentation methods that merely boost the diversity of the training samples, our method directly impacts the test samples by additional normalized images. However, these additional normalized images aim to reduce the diversity of task-irrelevant factors, thereby enhancing facial expression analysis performance.

\begin{table}[t] 
 \begin{minipage}[t]{0.42\textwidth}
\centering
 \setlength{\belowcaptionskip}{0.2cm}
\setlength{\abovecaptionskip}{-0cm} 
\tiny
\setlength{\arrayrulewidth}{0.3pt}
\begin{tablefont}
\setlength{\tabcolsep}{1.0mm}{
\caption{Comparison with data augmentation methods (\%).  Idn is replaced with these methods. 
}
\label{tab7}
\begin{tabular}{l|ccc|cc} 
\multirow{2}{*}{Method}  & \multicolumn{3}{c}{AU det(\textit{F1})}                                & \multicolumn{2}{|c}{FER(\textit{Acc.})}                                   \\
                       & \multicolumn{1}{c}{BP4D} & \multicolumn{1}{c}{BP4D+} & \multicolumn{1}{c}{DISFA} & \multicolumn{1}{|c}{RAF-DB} & \multicolumn{1}{c}{AffectNet}  \\ 
\specialrule{0.6pt}{0pt}{0pt}
Random init & 66.4 & 64.0 & 70.2 & 89.15  & 66.26\\
Mixup \cite{ref68}      & 67.2 & 64.8 & 70.8 & 89.24  & 67.29  \\
Cutout \cite{ref72}     & 66.7 & 64.2 & 70.3 & 89.28  & 66.49  \\
CutMix \cite{ref69}     & 67.4 & 64.5 & 70.4 & 89.57  & 66.74 \\ 
\specialrule{0.3pt}{0pt}{0pt}
Norface & \textbf{69.3} & \textbf{66.7} & \textbf{72.7} & \textbf{92.97}  & \textbf{68.69}  \\
\end{tabular}}
\end{tablefont}
\end{minipage}
\hfill
\begin{minipage}[t]{0.57\textwidth}
\centering
\tiny
\setlength{\arrayrulewidth}{0.3pt}
\setlength{\belowcaptionskip}{0.2cm}
\begin{tablefont}
\setlength{\tabcolsep}{0.52mm}{
\caption{Comparison with expression features (\%). Baseline means $w/o$ $E_{exp}\&E_{eye}$ and $w/o$ normalized images.}
\label{tabemb}
\begin{tabular}{l|ccc|cc} 
\multirow{2}{*}{Method}  & \multicolumn{3}{c}{AU det(\textit{F1})}                                & \multicolumn{2}{|c}{FER(\textit{Acc.})}                                  \\
                       & \multicolumn{1}{c}{BP4D} & \multicolumn{1}{c}{BP4D+} & \multicolumn{1}{c}{DISFA} & \multicolumn{1}{|c}{RAF-DB} & \multicolumn{1}{c}{AffectNet}  \\ 
\specialrule{0.6pt}{0pt}{0pt}
{\cellcolor[rgb]{0.902,0.902,0.902}}Baseline & {\cellcolor[rgb]{0.902,0.902,0.902}}67.2 & {\cellcolor[rgb]{0.902,0.902,0.902}}64.8 &{\cellcolor[rgb]{0.902,0.902,0.902}}70.2 & {\cellcolor[rgb]{0.902,0.902,0.902}}89.24  & {\cellcolor[rgb]{0.902,0.902,0.902}}67.29   \\
GLEE-Net \cite{ref62}  & 65.7  & 63.7 & 65.5 & - & - \\
GLEE-Net \cite{ref62} ($w/$ $E_{exp}\&E_{eye}$)  &65.6  & 63.9 & 65.8 & - & - \\
{\cellcolor[rgb]{0.902,0.902,0.902}}$w/ E_{exp}\&E_{eye}$ &{\cellcolor[rgb]{0.902,0.902,0.902}}67.1 ($\color{red}{\downarrow}$)  & {\cellcolor[rgb]{0.902,0.902,0.902}}64.6 ($\color{red}{\downarrow}$) &{\cellcolor[rgb]{0.902,0.902,0.902}}70.1 ($\color{red}{\downarrow}$) & {\cellcolor[rgb]{0.902,0.902,0.902}}89.77  & {\cellcolor[rgb]{0.902,0.902,0.902}}67.06 ($\color{red}{\downarrow}$) \\
\specialrule{0.3pt}{0pt}{0pt}
$w/$ Idn (Ours) & \textbf{69.3}  & \textbf{66.7}   & \textbf{72.7}& \textbf{92.97}    & \textbf{68.69}    \\
\end{tabular}}
\end{tablefont}
\end{minipage}
\end{table}

\noindent\textbf{A4. Images v.s. Expression features for classification.}
As mentioned before, our classification network takes normalized images as input. 
These images result from the Idn that is supervised by two types of expression feature, $E_{exp}\&E_{eye}$ for constraining the expression consistency. 
Intuitively, there exists another possibility to directly use $E_{exp}$\&$E_{eye}$ for classification, instead of normalized images. 
(1) Tab.\ref{tabemb} shows the results of $w/$ $E_{exp}\&E_{eye}$. 
In this setup, normalized images are omitted, and the used features are $w/$ $E_{exp}\&E_{eye}$ in replace of $e_{n_{id_*}}$ in the classification network. For a fair comparison, we maintain $e_{o_{'}}$. These features are fed into the Input and Output MoE modules.
Our method outperforms $w/$ $E_{exp}\&E_{eye}$ approach by 2.2\%, 2.1\%, 2.6\%, 3.20\%, and 1.63\% on the five datasets, respectively. 
(2) Moreover, our method is also compared with GLEE-Net\cite{ref62}, which is the SOTA method designed for AU detection using features from $E_{exp}$ and additional 3DMM. Tab.\ref{tabemb} shows that, on average, ours also performs better than GLEE-Net by 4.6\%.
(3) For a fair comparison, we replace GLEE-Net's features with $E_{exp}\&E_{eye}$, ours still leads by 4.5\%. 
The above results show the superiority of normalized images for classification.

It is interesting to discuss the disparities between normalized images and expression features. Compared to highly abstract expression features, images encompass more structured details at pixel level. Furthermore, the used MAE can refine the facial representation from those normalized images due to its training on large-scale datasets. These factors contribute to the superior performance of normalized images over expression features.






\subsection{Ablation Study}
Ablation studies are conducted to validate expression consistency loss in the normalization network and each component in the classification Network.

\begin{figure}[t]
\begin{minipage}[b]{.45\linewidth}
\centering
\tiny
\setlength{\arrayrulewidth}{0.3pt}
\begin{tablefont}
\setlength{\tabcolsep}{0.8mm}{
\captionof{table}{Results of AU detection $w/$ and $w/o$ $\mathcal{L}_{eye}$ in Eyebrow Region on BP4D in terms of F1 scores (\%).}
\label{tabloss}
\setlength{\abovecaptionskip}{-0.1cm} 
\setlength{\belowcaptionskip}{-0cm}
\begin{tabular}{c|ccccc|c}
Losses & AU1  & AU2  & AU4  & AU6  & AU7  & Avg.  \\
\specialrule{0.6pt}{0pt}{0pt}
$w/$ $\mathcal{L}_{exp}$      & 60.2 & 54.9 & 67.1 & 79.8 & 79.8 & 68.3  \\
\textbf{$w/$ $\mathcal{L}_{exp} \& \mathcal{L}_{eye}$}     &  \textbf{60.7} &  \textbf{55.1} &  \textbf{67.7} &  \textbf{79.9} &  \textbf{80.4} & \textbf{68.8}  \\
\end{tabular}}
\end{tablefont}
\end{minipage}%
  \hfill
\begin{minipage}[b]{.5\linewidth}
\centering
\setlength{\abovecaptionskip}{0cm}
\includegraphics[scale=0.21]{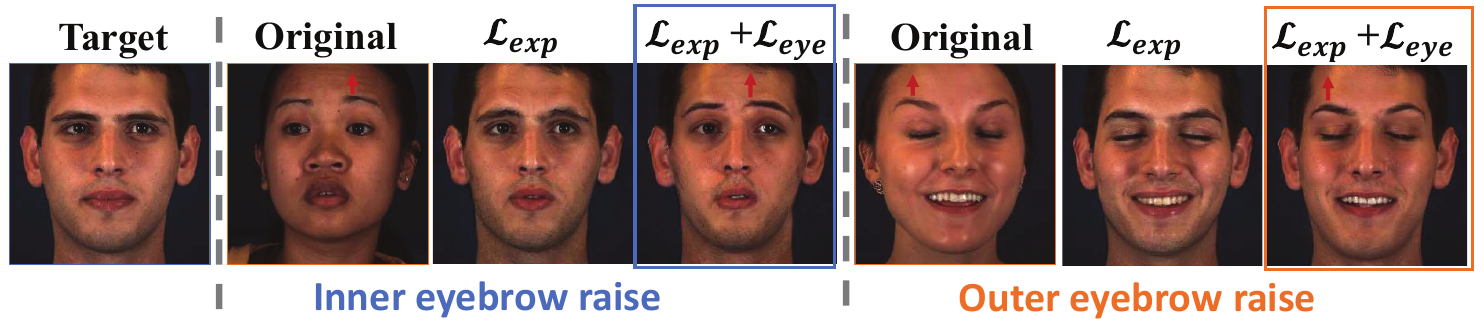}
\caption{Visualization of the effects of the expression loss $\mathcal{L}_{exp}$ and eyebrow loss $\mathcal{L}_{eye}$ in normalization network.} 
\label{fig_aba}
\end{minipage}%
\end{figure}

\noindent\textbf{Expression consistency loss.} 
The expression loss $\mathcal{L}_{exp}$ and eyebrow loss $\mathcal{L}_{eye}$ in $F^{nor}$ are used to ensure expression consistency. 
Tab.\ref{tabloss} validates their effectiveness, presenting the results for $w/$ both $\mathcal{L}_{exp}$ and $\mathcal{L}_{eye}$, as well as only $w/$ $\mathcal{L}_{exp}$ in the AU detection task. The result shows that, in the eyebrow regions, using both $\mathcal{L}_{exp}$ and $\mathcal{L}_{eye}$ yields an average 0.5\% higher than merely using $\mathcal{L}_{exp}$. While $\mathcal{L}_{exp}$ guides the full face, $\mathcal{L}_{eye}$ enhances the eyebrow regions. Moreover, without $\mathcal{L}_{exp}$, Idn would fail, as $\mathcal{L}_{eye}$ solely concentrates on the eyebrow regions.
Fig.\ref{fig_aba} further visualizes their impacts by a few samples. 

\begin{figure}[t]
\centering
\begin{minipage}[t]{.48\textwidth}
\centering
\includegraphics[scale=0.45]{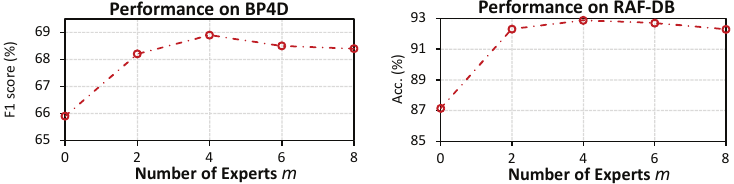}
\caption{Comparison of different numbers of experts ${m}$.}
\label{fig_moe_af}
\end{minipage}
\hfill
\begin{minipage}[t]{.48\textwidth}
\centering
\includegraphics[scale=0.2]{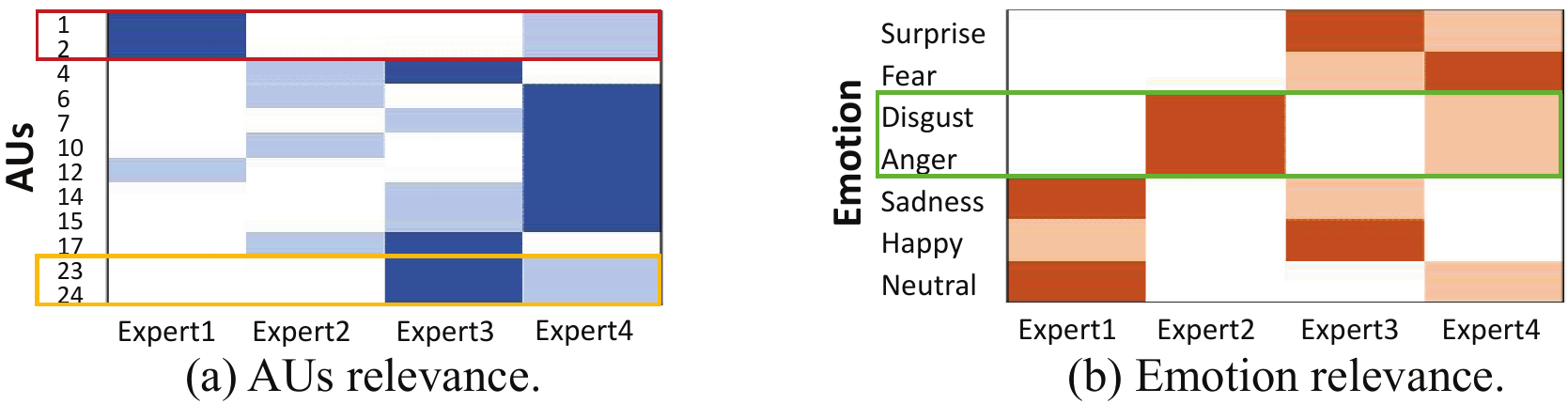}
\caption{Router decision statistics for the Output MoE module with Top-$2$ on both tasks. Dark colors represent high weight. }
\label{fig_moe}
\end{minipage}
\end{figure}

\noindent\textbf{Classification Network.} 
Fig.\ref{fig_moe_af} shows the results of different numbers of experts $m$ under AU detection and FER tasks.  As $m$ increases, the performance of MoE improves, peaking at $m=4$ with a performance boost of 3.0\% and 5.74\% compared to $m=0$.
In addition, Tab.\ref{tabmoe} reports the results $w/$ and $w/o$ the Input MoE module $M_{i}$ and the Output MoE module $M_{o}$, demonstrating the contribution of each module.
Furthermore, Fig.\ref{fig_moe}  visualizes the statistics of the Top-$2$ experts selected by the Output MoE module. 
For example, in the AU detection task, Expert $1$ and Expert $4$ primarily focus on detecting AU1 and AU2. These results imply that the Output MoE module effectively focuses on specific tasks, thus enhancing task-related inductive learning. Tab.\ref{tab_loss} shows the effectiveness of each loss in $F^{cla}$.

\begin{table}[htbp] 
 \begin{minipage}[t]{0.42\textwidth}
\centering
\tiny
\setlength{\arrayrulewidth}{0.3pt}
\setlength{\abovecaptionskip}{0cm} 
\setlength{\belowcaptionskip}{-0.01cm}
\begin{tablefont}
\setlength{\tabcolsep}{0.1mm}{
\caption{Ablation study with MoE modules (\%). 
}
\label{tabmoe}
\begin{tabular}{c|c|c|c|c}
Strategy & $w/o\ M_i\ \&\ M_o$  & $w/o\ M_i$  & $w/o\ M_o$  & Ours  \\
\specialrule{0.6pt}{0pt}{0pt}
BP4D (\textit{F1})       & 65.9 & 68.4 & 68.8 & \textbf{69.3}  \\
RAF-DB  (\textit{Acc.})       & 87.15 & 91.92 & 92.57 & \textbf{92.97}  \\
\end{tabular}}
\end{tablefont}
  \end{minipage}
  \hfill
   \begin{minipage}[t]{0.5\textwidth}
   \centering
\tiny
\setlength{\arrayrulewidth}{0.3pt}
\begin{tablefont}
\setlength{\abovecaptionskip}{-0.02cm} 
\setlength{\belowcaptionskip}{-0.01cm}
\setlength{\tabcolsep}{0.1mm}{
\caption{Ablation study with different losses (\%).}
\label{tab_loss}
\begin{tabular}{c|c|c|c|c}
Losses & $w/o\ \mathcal{L}_{imp}$  & $w/o\ \mathcal{L}_{global}$  & $w/o\ \mathcal{L}_{local}$  & Ours  \\
\specialrule{0.6pt}{0pt}{0pt}
BP4D (\textit{F1})       & 68.5 & 68.7 & 68.9 & \textbf{69.3}  \\
RAF-DB  (\textit{Acc.})      & 91.66 & 92.08 & 92.57 & \textbf{92.97}  \\
\end{tabular}}
\end{tablefont}
 \end{minipage}
\end{table}




\section{Conclusion}
This paper provides a novel insight and a unified framework called Norface for AU detection, AU intensity estimation, and FER tasks, aiming to remove the task-irrelevant noise by identity normalization that normalizes all images to a common identity with consistent pose and background, and improves expression classification. Extensive quantitative and qualitative experiments demonstrate the superior performance of Norface for AU analysis and FER tasks as well as their cross-dataset tasks. 

%
%
\bibliographystyle{splncs04}
\bibliography{reference}

%
%
\newpage 

\section*{Supplementary Material} 

\begin{figure*}[t] 
\centering 
\includegraphics[width=0.9\textwidth]{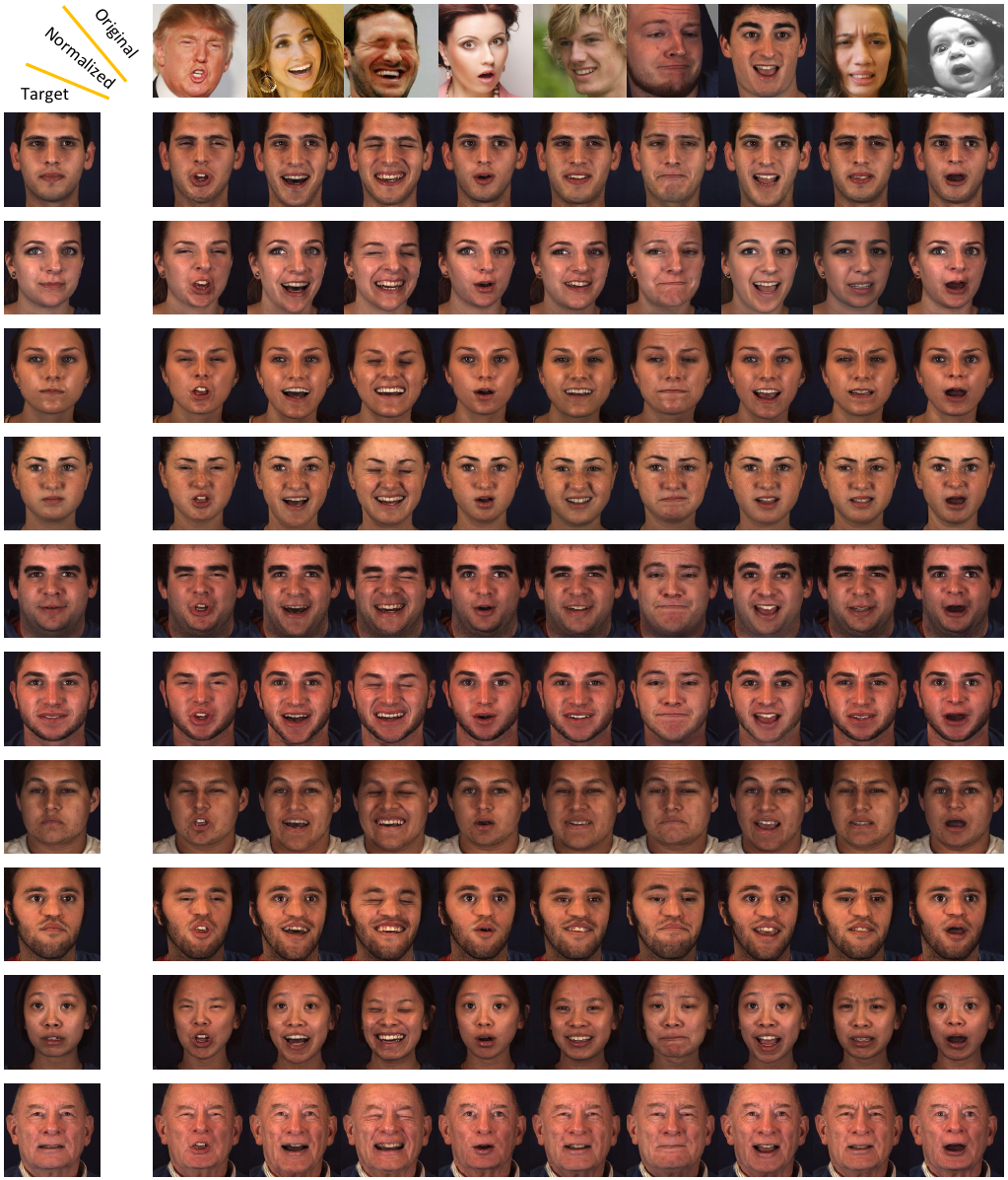}
\caption{Identity normalization with various target faces. Please zoom in for more details.} 
\label{fig22} 
\end{figure*}

\begin{figure*}[t] 
\centering 
\includegraphics[width=0.82\textwidth]{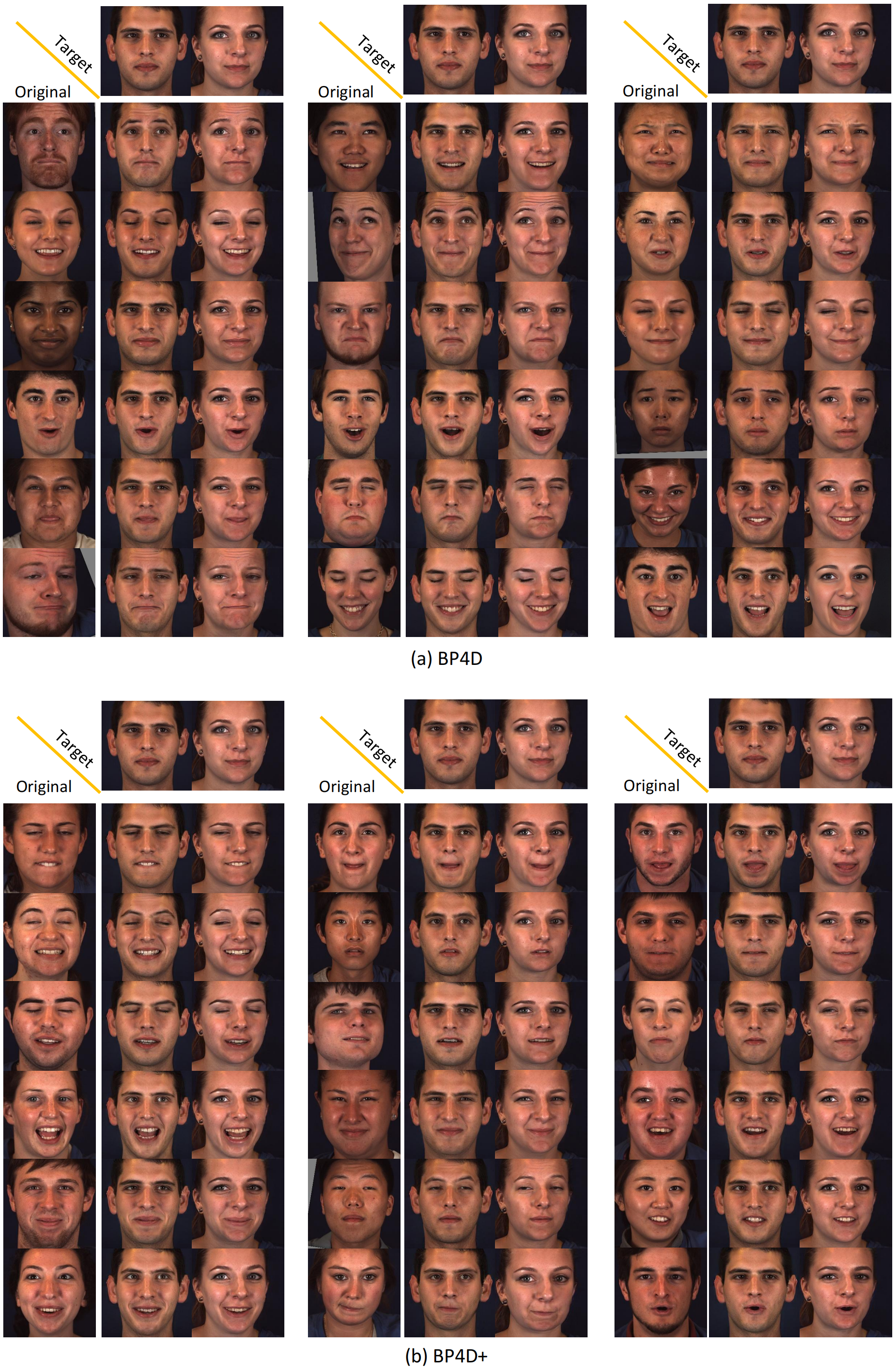}
\caption{Identity normalization of BP4D and BPD4+ datasets as original faces. The figure presents results on target faces with two identities. Please zoom in for more details.} 
\label{fig33} 
\vspace{-0.5em}
\end{figure*}

\begin{figure*}[t] 
\centering 
\includegraphics[width=0.82\textwidth]{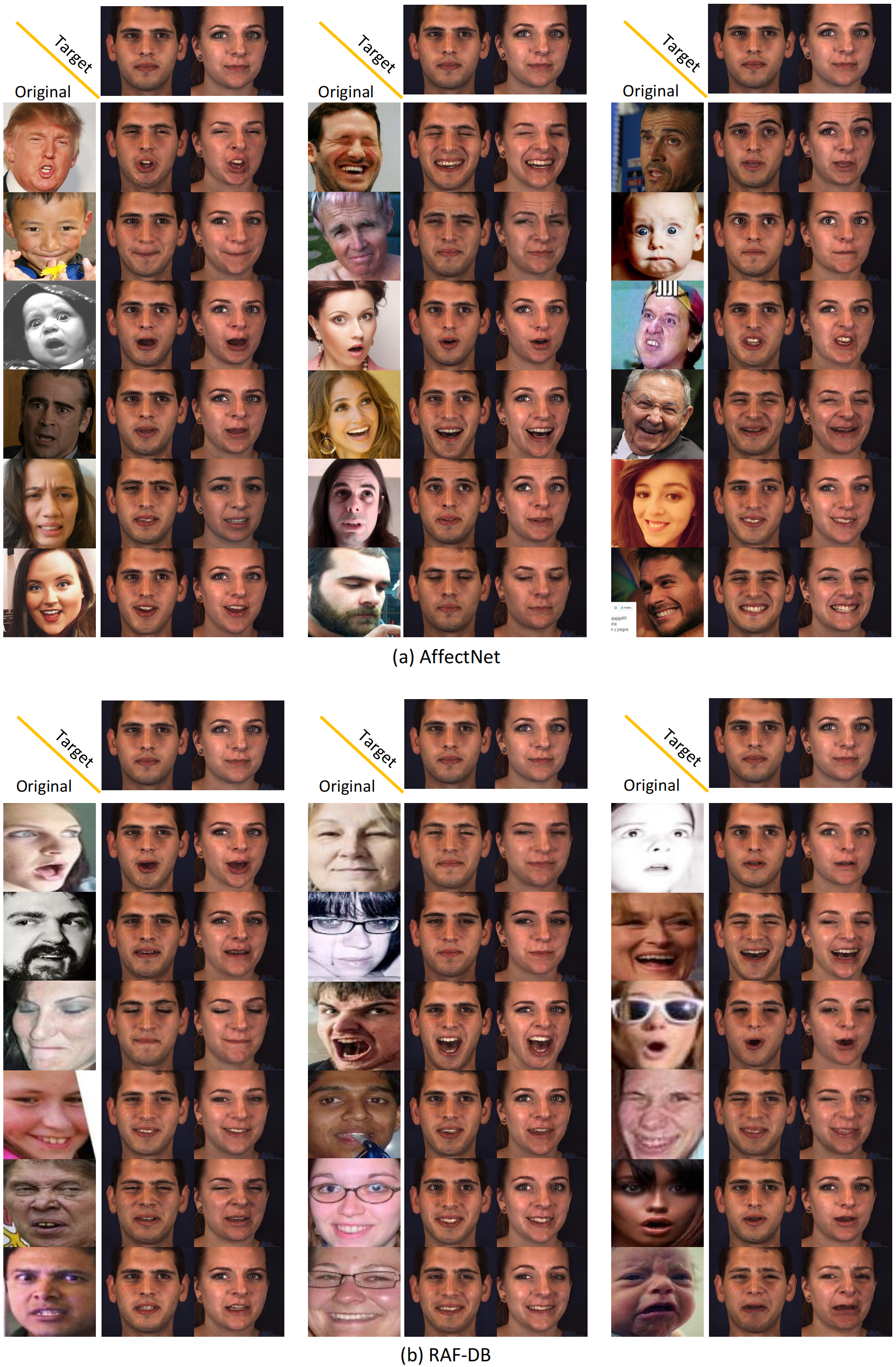}
\caption{Identity normalization of AffectNet and RAF-DB datasets as original faces. The figure presents results on target faces with two identities. Please zoom in for more details.} 
\label{fig44} 
\vspace{-0.5em}
\end{figure*}

\begin{figure*}[t] 
\centering 
\includegraphics[width=0.82\textwidth]{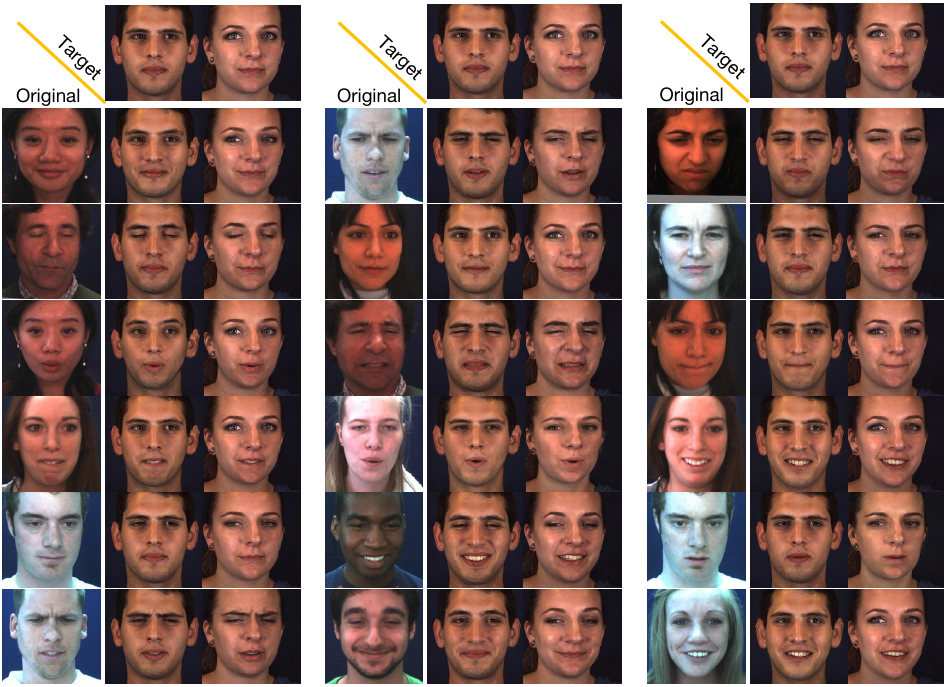}
\caption{Identity normalization of DISFA dataset as original faces. The figure presents results on target faces with two identities. Please zoom in for more details.} 
\label{fig55} 
\vspace{-0.5em}
\end{figure*}

Fig.\ref{fig22} showcases several original images that have been normalized to multiple target identities while maintaining consistency in facial expressions. Furthermore, Fig.\ref{fig33}, Fig.\ref{fig44}, and Fig.\ref{fig55} show more normalization results from BP4D \cite{ref1}, BPD4+ \cite{ref2}, AffectNet \cite{ref3}, RAF-DB \cite{ref4}, and DISFA \cite{ref5} datasets that are utilized in our study. The visualization results demonstrate the versatility of our normalization network.

We will release all the images normalized from BP4D \cite{ref1}, BPD4+ \cite{ref2}, AffectNet \cite{ref3}, RAF-DB \cite{ref4}, and DISFA \cite{ref5} datasets. We believe that these normalized images will help other research works on facial expression classification or analysis.

\end{document}